\documentclass[journal]{IEEEtran}

\ifCLASSINFOpdf
\else
\fi

\usepackage{times}
\usepackage{epsfig}
\usepackage{graphicx}
\usepackage{amsmath}
\usepackage{amssymb}
\usepackage{mathrsfs}
\usepackage{multirow}
\usepackage{blkarray}
\usepackage{float}
\usepackage{array}
\usepackage{epstopdf}
\usepackage{subcaption}
\usepackage[T1]{fontenc}
\usepackage{algorithm}
\usepackage{algpseudocode}

\begin{document}

\title{Image Segmentation Using Overlapping Group Sparsity}

\author{Shervin~Minaee,~\IEEEmembership{Student Member,~IEEE,}
        and~Yao~Wang,~\IEEEmembership{Fellow,~IEEE}
}

\maketitle

\begin{abstract}
Sparse decomposition has been widely used for different applications, such as source separation, image classification and image denoising. 
This paper presents a new algorithm for segmentation of an image into background and foreground text and graphics using sparse decomposition.
First, the background is represented using a suitable smooth model, which is a linear combination of a few smoothly varying basis functions, and the foreground text and graphics are modeled as a sparse component overlaid on the smooth background.
Then the background and foreground are separated using a sparse decomposition framework and imposing some prior information, which promote the smoothness of background, and the sparsity and connectivity of foreground pixels.
This algorithm has been tested on a dataset of images extracted from HEVC standard test sequences for screen content coding, and is shown to outperform prior methods, including least absolute deviation fitting, k-means clustering based segmentation in DjVu, and shape primitive extraction and coding algorithm.
\end{abstract}


\IEEEpeerreviewmaketitle

\section{Introduction}
Image segmentation can be thought as a signal decomposition problem, where the goal is to decompose the signal into several components such that each one represents the same content or semantic information.
One special case is the background-foreground segmentation, which tries to decompose an image into two components, background and foreground.
Foreground-background segmentation has many applications in image processing such as separate coding of background and foreground in video compression, organ detection in medical images, text extraction, and medical image analysis \cite{MRC}-\cite{chromosome}.

Different algorithms have been proposed in the past for foreground-background segmentation, including k-means clustering in DjVu \cite{djvu}, shape primitive extraction and coding (SPEC) \cite{spec}, least absolute deviation fitting \cite{LAD}, and sparse-smooth decomposition \cite{SPSD}.
The hierarchical k-means clustering applies the k-means clustering algorithm with k=2 on blocks in multi-resolution. It first applies the k-means clustering algorithm on a large block to obtain foreground and background colors and then uses them as the initial foreground and background colors for the smaller blocks in the next stages. It also applies some post-processing at the end to refine the results. This algorithm has difficulty for the regions where background and foreground color intensities overlap and some part of the background will be detected as foreground.
In the shape primitive extraction and coding (SPEC) method, a two-step segmentation algorithm is proposed. 
In the first step the algorithm classifies each block of size $16 \times 16$ into either pictorial block or text/graphics, by comparing the number of colors with the threshold 32.
In the second step, segmentation result of pictorial blocks is refined by extracting shape primitives.
Because of variations in image content, it is hard to find a fixed threshold on the number of colors that can robustly separate pictorial blocks and text/graphics blocks.

One problem with clustering-based segmentation techniques is that if the intensity of background pixels has a large dynamic range, some part of the background could be segmented as foreground.
One such example is shown in Fig. 1, where the foreground mask (a binary mask showing the location of foreground pixels) for a sample image by hierarchical clustering and the proposed algorithm are shown.
\begin{figure}[1 h]
\begin{center}
\hspace{-0.15cm}
    \includegraphics [scale=0.44] {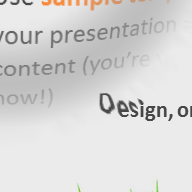}
  \hspace{0.11cm}  \includegraphics [scale=0.26] {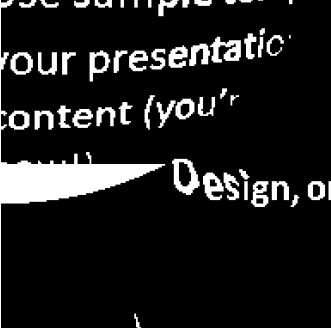} 
  \hspace{-0.29cm}  \includegraphics [scale=0.15] {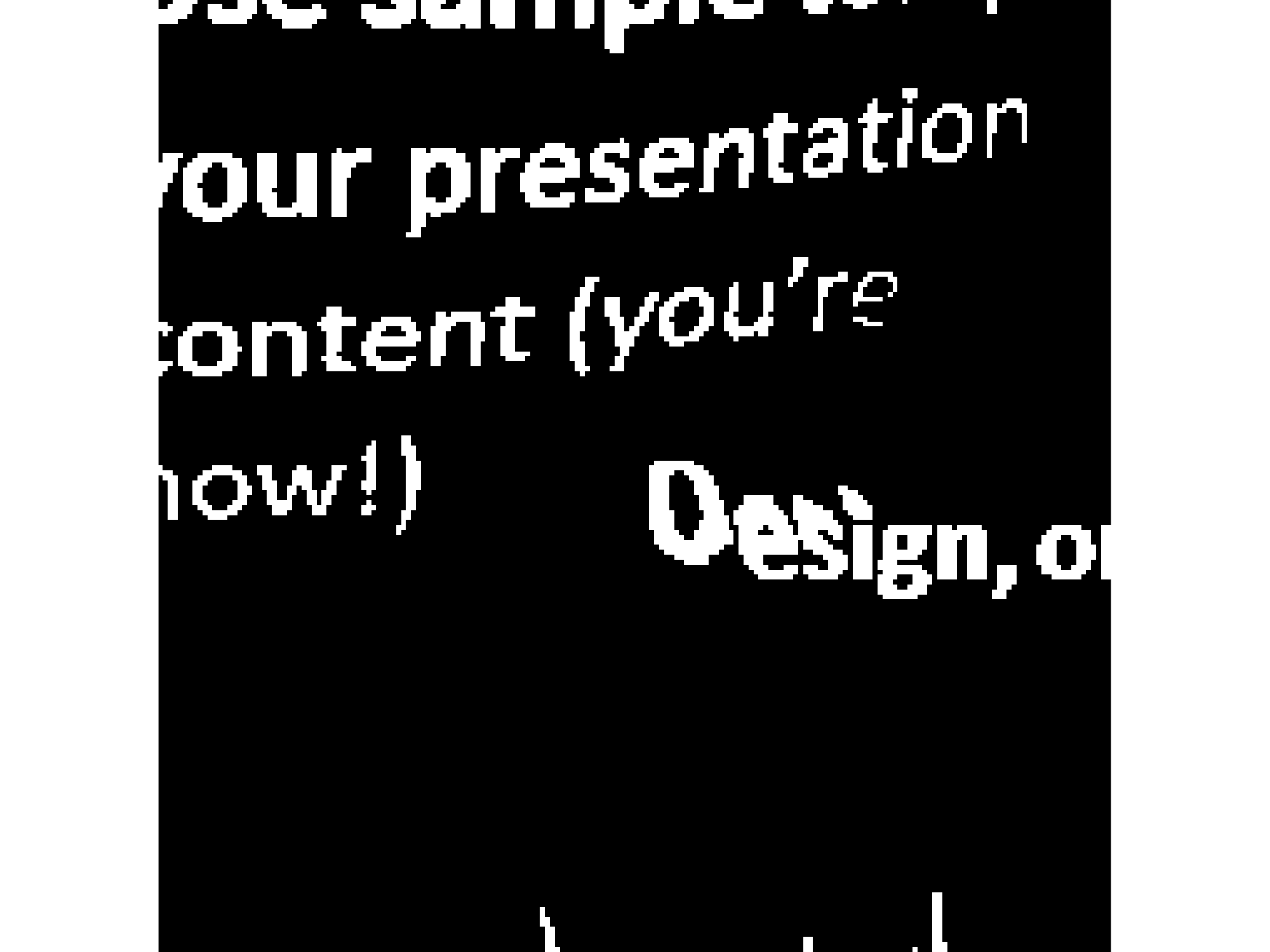} 
\end{center}
  \caption{The left, middle and right images denote the original image, segmented foreground by hierarchical k-means and the proposed algorithm respectively.  }
\end{figure}
\vspace{-0.1cm}

To overcome this problem, in a previous work we proposed a least absolute deviation fitting method, which fits a smooth model to the image and classifies regions in the image based on their smoothness \cite{LAD}. 
It uses the $\ell_1$ norm on the fitting error to enforce the sparsity of the error term. Although this algorithm achieved significantly better segmentation than both DjVu and SPEC, it suffers from two problems.
First, it does not impose any connectivity on the pixels in the foreground layer, which could result in many isolated points in the foreground. 
Second, it uses a fixed set of smooth basis functions and does not impose any restrictions on the smooth model parameters, which could result in overfitting the smooth model to the image.

In this work we propose a sparse decomposition based segmentation, which tries to resolve these problems.
Sparse representation has been used for various applications in recent years, including face recognition , super-resolution, morphological component analysis, image restoration, image denoising and sparse coding \cite{wright}-\cite{sp_coding2}.
Within our sparse decomposition framework, we also impose suitable priors on each layer, in particular smoothness on the background, and connectivity on the foreground.
To promote the connectivity of the foreground component, the group-sparsity of the foreground pixels is added to the cost function. It is worth mentioning that total-variation can also be used to promote connectivity, and it is used in some of our previous works \cite{seg_tv}-\cite{seg_journal}.
This algorithm is tested on an image segmentation dataset and is shown to be very successful.

The structure of the rest of this paper is as follows: Section II presents the main idea of the proposed segmentation algorithm. Section III describes the ADMM formulation for solving the proposed optimization problem. Section IV provides the experimental results for the proposed algorithm, and the paper is concluded in Section V.

\section{Sparse Decomposition Framework}
The proposed sparse decomposition algorithm in this work segments a given image into two layers, background and foreground. 
The background contains the smooth part of the image and can be well represented with a few smooth basis functions,
whereas the foreground contains the text and graphics, which cannot be represented with a smooth model. 
Using the fact that foreground pixels usually occupy a small percentage of the images, we can model them with a sparse component overlaid on top of the background. 
Therefore it makes sense to think of mixed content image as a superposition of two layers, one smooth and the other one sparse. Hence we can use sparse decomposition techniques to separate these two components.

Each image is first divided into non-overlapping blocks of size $N\times N$, denoted with $F(x,y)$, where $x$ and $y$ designate the horizontal and vertical axes. 
Then it is represented as a sum of two components $F= X+S$, where $X$ and $S$ denote the smooth background and sparse foreground components respectively. 
The background is modeled as a linear combination of $K$ basis functions $\sum_{k=1}^K \alpha_k P_k(x,y)$, where $P_k(x,y)$ denotes a 2D smooth basis function \cite{LAD}, and $\alpha_1,...,\alpha_K$ denote the parameters of this smooth model.
Since this model is  linear in parameters, $\alpha_k$, it is simpler to find the optimal weights.
To find $P_k(x,y)$, the Karhunen-Loeve transform \cite{KLT} is applied to a training set of smooth background images and it turns out the optimal learned bases are very similar to 2D DCT bases.
Therefore a set of low-frequency two-dimensional DCT basis functions are used in our model.
DCT bases are shown to be very efficient for image representation and compression \cite{DCT}. 
All the possible basis functions are ordered in the conventional zig-zag order in the $(u,v)$ plane, and the first $K$ basis functions are chosen. 
Hence each image block can be represented as:
\begin{equation}
F(x,y)= \sum_{k=1}^K \alpha_k P_k(x,y) + S(x,y)
\end{equation}
where $\sum_{i=1}^K \alpha_i P_i(x,y)$ and $S(x,y)$ correspond to the smooth background region and foreground pixels respectively. 
\\To have a more compact notation, we convert all 2D blocks of size $N \times N$ into vectors of length $N^2$, denoted by $f$ and $s$ (for $F(x,y)$ and $S(x,y)$), and show $\sum_{k=1}^K \alpha_k P_k(x,y)$ as $ {P\alpha}$ where ${P}$ is a matrix of size $N^2\times K$ in which the $k$-th column corresponds to the vectorized version of $P_k(x,y)$ and $\alpha=[\alpha_1,...,\alpha_K]^\text{T}$.
Then Eq. (1) can be written as:
\begin{equation}
f= {P\alpha}+s
\end{equation}

The decomposition problem in Eq. (2) is a highly ill-posed problem. 
Therefore we need to impose some priors on $\alpha$ and $s$ to be able to perform this decomposition.
Here three priors are imposed, sparsity of $\alpha$, sparsity of the foreground and connectivity of the foreground.
The reason for imposing sparsity on $\alpha$ is that we do not want to use too many basis functions for background representation.
Without such a restriction on the coefficients, we might end up with the situation in which all foreground pixels are also modeled by the smooth layer.
The second prior, sparsity of foreground, is inspired from the fact that the foreground pixels are expected to occupy a small percentage of pixels in each block.
The last but not the least point is that we expect the nonzero components of the foreground to be connected to each other and we do not want to have a set of isolated points detected as foreground. 
Therefore we can add a group sparsity regularization to promote the connectivity of the foreground pixels.
\\
We can incorporate all these priors in an optimization problem as shown below:
\begin{equation}
\begin{aligned}
& \underset{s, \alpha}{\text{minimize}}
& & \|\alpha \|_0+ \lambda_1 \| s \|_0+ \lambda_2 G(s)   \\
& \text{subject to}
& &  f=P \alpha+s
\end{aligned}
\end{equation}
where $\lambda_1$ and $\lambda_2$ are the weights for regularization terms that need to be tuned, and $G(s)$ shows the group-sparsity on foreground.
For the first two terms since $\ell_0$ is not convex, we replace them with $\ell_1$ which is its relaxed-convex version.
For the group-sparsity we used the mixed  $\ell_{1}/\ell_2$ norm of $s$ \cite{bach1}, which is defined as:
\begin{equation}
\begin{aligned}
G(s)=  \sum_{i} ||s_{g_i}||_2
\end{aligned}
\end{equation}
where $g_i$ denotes the $i$-th group. Here we have overlapping groups, which consist of all columns and rows in the image. 
Therefore we can show the group sparsity term as the summation of two terms, one over all columns and the other one over all rows of image as:
\begin{equation}
\begin{aligned}
G(s)=  \sum_{i \in \text{rows}} ||s_{r_i}||_2+ \sum_{j \in \text{columns}} ||s_{c_j}||_2
\end{aligned}
\end{equation}
Then we will get the following problem:
\begin{equation}
\begin{aligned}
& \underset{s, \alpha}{\text{minimize}}
& & \|\alpha \|_1+ \lambda_1 \| s \|_1+ \lambda_2 \big( \sum_{i} ||s_{r_i}||_2+ \sum_{j} ||s_{c_j}||_2  \big)   \\
& \text{subject to}
& &  P \alpha+s=f
\end{aligned}
\end{equation}

This problem can be solved with different approaches, such as proximal optimization \cite{patrick1}, alternating direction method of multipliers (ADMM) \cite{ADMM}, and SLASSO algorithm \cite{SLASSO}. 
Here we present the formulation using ADMM algorithm.

\section{ADMM algorithm for the proposed sparse decomposition}
ADMM is a popular algorithm, which combines the superior convergence properties of method of multiplier and decomposability of dual ascent. 
The ADMM formulation for the optimization problem in (6) can be derived by introducing auxiliary variables as:
\begin{equation}
\begin{aligned}
& \underset{\alpha, \beta, s, y, z}{\text{minimize}}
& & \| \beta \|_1+ \lambda_1 \| s \|_1+ \lambda_2 \big( \sum_{i} ||y_{r_i}||_2+ \sum_{j} ||z_{c_j}||_2  \big) \\
& \text{subject to} & &  f= P \alpha+s \\
& & & \alpha= \beta, s= y, \ s=z 
\end{aligned}
\end{equation}
Then the augmented Lagrangian for the above problem can be formed as:
\begin{align*}
&L_{\rho_1,\rho_2, \rho_3, \rho_4}(\alpha, s, \beta, y, z, w_1, w_2, v_1, v_2)= \| \beta \|_1+ \lambda_1 \| s \|_1+ 
\nonumber \\ &\lambda_2 \big( \sum_{i=1}^N ||y_{r_i}||_2+ \sum_{j=1}^N ||z_{c_j}||_2  \big)+ w_1^t(f- P\alpha-s)+  \nonumber \\ &w_2^t(\alpha-\beta)+ v_1^t (s-y)+ v_2^t(s-z)
 +\frac{\rho_1}{2} \| f- P\alpha-s \|_2^2+ \nonumber \\
 &\frac{\rho_2}{2} \| \alpha- \beta \|_2^2+ \frac{\rho_3}{2} \| s-y \|_2^2+ \frac{\rho_4}{2} \| s-z \|_2^2
\end{align*}
where $w_1$, $w_2$, $v_1$ and $v_2$ denote the dual variables, and $\rho_{1:4}$ denote the penalty terms.

Then, we can solve this optimization problem by taking the gradient of the objective function w.r.t. the primal variables and setting it to zero and using dual descent for dual variables. After doing some simplification, we will get the update rule described in Algorithm 1 for solving the optimization problem in (7).
\begin{algorithm}
  \caption{pseudo-code for ADMM updates of problem (7)}\label{euclid}
  \begin{algorithmic}[1]
      \For{\texttt{$k$=1:$k_{max}$}} \vspace{0.06cm} 
        \State $\alpha^{k+1}= A^{-1} \big[ P^t w_1^k- w_2^k + \rho_2 \beta^k+\rho_1 P^T (f-s^k) \big]$ \vspace{0.15cm}
        \State $\beta^{k+1}= \text{Soft}(\alpha^{k+1}+ \frac{w_2^k}{\rho_2},\frac{1}{\rho_2}) $   \vspace{0.15cm}     
        \State $s^{k+1}= \frac{1}{\rho_1+\rho_3+\rho_4} \text{Soft}( C, \lambda_1) $  \vspace{0.15cm}
            \For{\texttt{$i$=1:$N$}} \vspace{0.06cm} 
            \State $y_{r_i}^{k+1}= \text{Block-Soft}( s_{r_i}^{k+1}+ \frac{v_{1,r_i}^k}{\rho_3} , \frac{\lambda_2}{\rho_3})$ \vspace{0.06cm} 
            \EndFor \vspace{0.15cm} 
            \For{\texttt{$j$=1:$N$}} \vspace{0.06cm} 
            \State $z_{c_j}^{k+1}= \text{Block-Soft}( s_{c_j}^{k+1}+ \frac{v_{2,c_j}^k}{\rho_3} , \frac{\lambda_2}{\rho_4})$ \vspace{0.06cm} 
            \EndFor \vspace{0.15cm} 
        \State $w_1^{k+1}= w_1^k+ \rho_1 (f- P\alpha^{k+1}-s^{k+1})  $       \vspace{0.15cm} 
        \State $w_2^{k+1}= w_2^k+ \rho_2 (\alpha^{k+1}- \beta^{k+1}) $ \vspace{0.15cm}
        \State $v_1^{k+1}= v_1^k+ \rho_3 (s^{k+1}-y^{k+1})  $  \vspace{0.15cm}
        \State $v_2^{k+1}= v_2^k+ \rho_4 (s^{k+1}-z^{k+1})  $          \vspace{0.15cm}     
      \EndFor
  \end{algorithmic}
\end{algorithm}

Here $A=(\rho_1 P^tP+\rho_2 I)$, $C= w_1^{k}-v_1^{k}-v_2^{k}+\rho_1 (f- P\alpha^{k+1})+  \rho_3 y^{k}+\rho_4 z^{k}$, and $\text{Soft}(x,\lambda)$ denotes the soft-thresholding operator \cite{soft} applied element-wise and is defined as:
\begin{gather*}
\text{Soft}(x,\lambda)= \text{sign}(x) \ \text{max}(|x|-\lambda,0)
\end{gather*}
and $\text{Block-Soft}(x,\lambda)$ is the soft-thresholding operator defined for vectors as:
\begin{gather*}
\text{Block-Soft}(x,\lambda)= (1- \lambda/{\|x\|_2})_+ x
\end{gather*}
with $\text{Block-Soft}(0,\lambda)=0$. For scalar $X$, block soft-thresholding reduces to the scalar soft-thresholding operator.

\section{Experimental Results}
To evaluate the performance of our algorithm, we have tested the proposed algorithm on the same screen content image segmentation dataset provided in \cite{LAD}, which consist of 332 image blocks of size 64x64, extracted from sample frames of HEVC test sequences for screen content coding \cite{SCC_data}, \cite{SCC_tran}.

In our implementation, the block size is chosen to be $N$=64. 
The number of DCT basis functions, $K$, is chosen to be 10. 
The weight parameters in the objective function are tuned by testing on a validation set and are set to be $\lambda_1=100$ and $\lambda_2= 2$. 
The ADMM algorithm described in Algorithm 1 is implemented in MATLAB and is publicly available in \cite{our_dataset}.
The number of iterations for ADMM is chosen to be 50 and the parameters $\rho_{1:4}$ are all set to 1.

We compare the proposed algorithm with three previous algorithms; hierarchical k-means clustering in DjVu, SPEC, and LAD. For SPEC, we have adapted the color number threshold and the shape primitive size threshold from the default value given in \cite{spec}  when necessary to give a more satisfactory result.

To provide a numerical comparison, we report the average precision, recall and F1 score \cite{metrics} achieved by different algorithms over this dataset. The average precision, recall and F1 score by different algorithms are given in Table 1. 

\begin{table}[ht]
\centering
  \caption{Comparison of accuracy of different algorithms}
  \centering
\begin{tabular}{|m{3.4cm}|m{1.2cm}|m{1.2cm}|m{1.2cm}|}
\hline
Segmentation Algorithm  &  \  \ Precision & \ \  Recall & \  F1 score\\
\hline
SPEC \cite{spec} & \ \ \ 50\% & \ \ \  64\% & \ \ \ 56.1\% \\
\hline
 Hierarchical Clustering \cite{djvu} & \ \ \ 64\% & \ \ \ 69\% & \ \ \ 66.4\% \\
\hline 
 Least Absolute Deviation \cite{LAD} & \ \ \  91.4\% & \ \ \  87\% & \ \ \  89.1\% \\
\hline
 The proposed algorithm & \ \ \ 93.7\%  & \ \ \ 86.7\%  & \ \ \  90\%\\
\hline
\end{tabular}
\label{TblComp}
\end{table}

The precision and recall are defined as in Eq. (8), where TP, FP and FN denote true positive, false positive and false negative respectively. In our evaluation, we treat a foreground pixel as positive. A pixel that is correctly identified as foreground (compared to the manual segmentation) is considered true positive. The same holds for false negative and false positive. The balanced F1 score is defined as the harmonic mean of precision and recall, as it is shown in Eq 9.
\begin{gather}
 \text{Precision}= \frac{\text{TP}}{\text{TP+FP}} \ , 
\ \ \ \ \text{Recall}= \frac{\text{TP}}{\text{TP+FN}} 
\end{gather}
\begin{gather}
\text{F1}= 2 \ \frac{\text{precision} \times \text{recall}}{\text{precision+recall}}
\end{gather}

\begin{figure*}
        \centering
        \vspace{-0.5cm}
        \begin{subfigure}[b]{0.18\textwidth}
                \includegraphics[width=\textwidth]{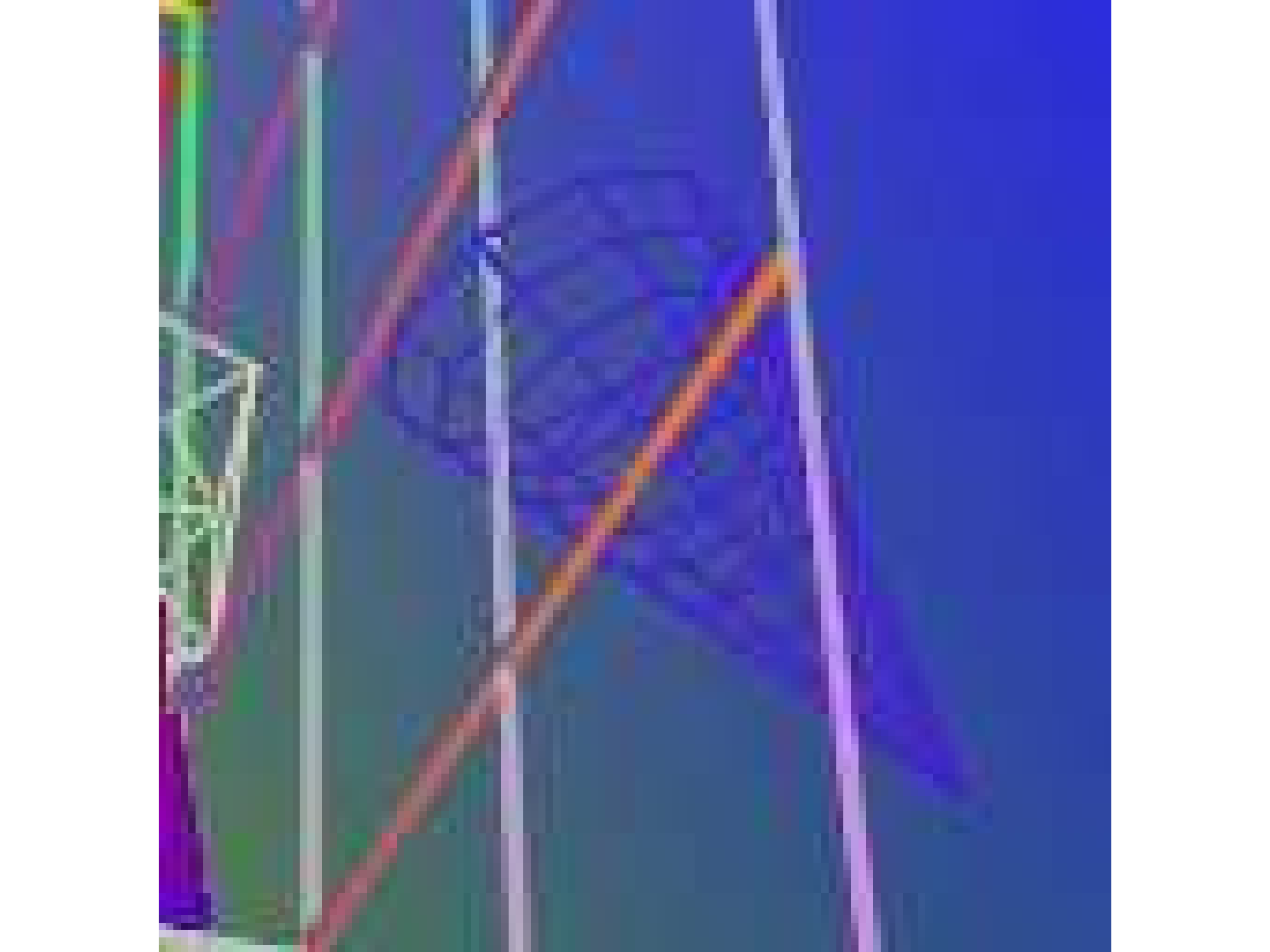}
                                \vspace{-0.5cm}
          \hspace{-2.5cm}    
        \end{subfigure}%
        ~ 
        \begin{subfigure}[b]{0.18\textwidth}
                \includegraphics[width=\textwidth]{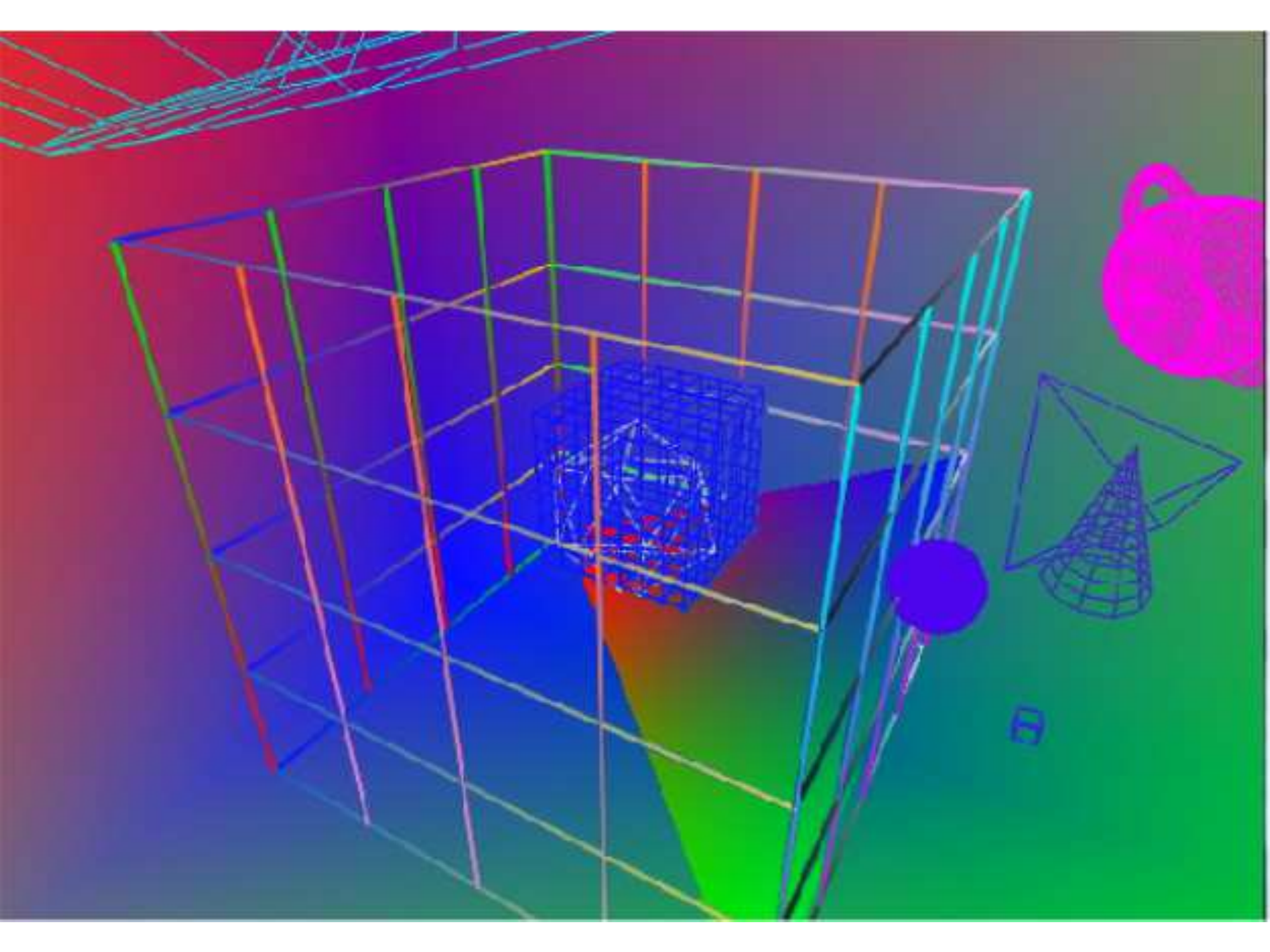}
                \vspace{-0.5cm}
            \hspace{-3cm} 
        \end{subfigure}%
        ~ 
        \begin{subfigure}[b]{0.18\textwidth}
                \includegraphics[width=\textwidth]{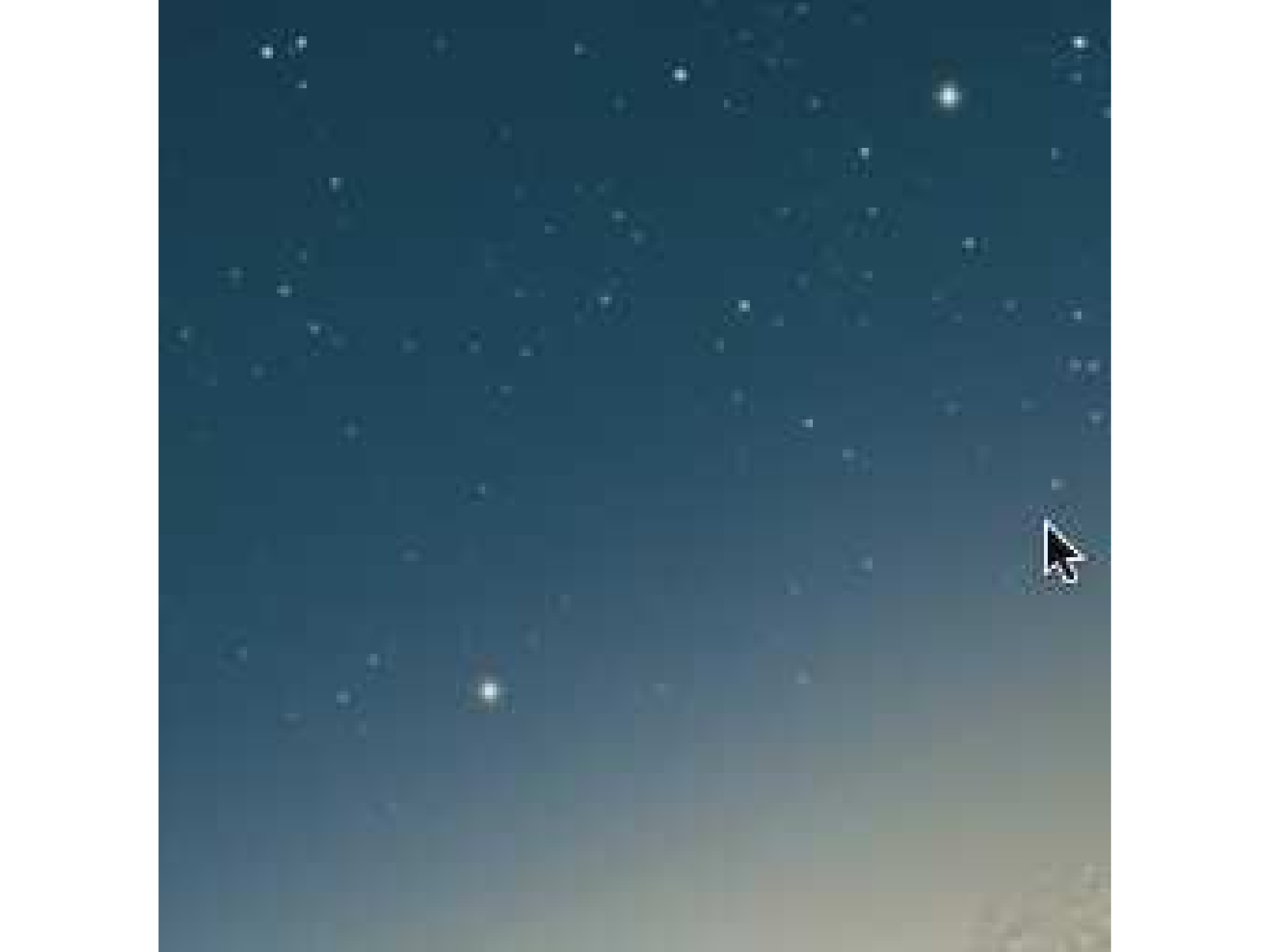}
                \vspace{-0.45cm}
            \hspace{-3cm} 
        \end{subfigure}%
        \begin{subfigure}[b]{0.18\textwidth}
			~ 
                \includegraphics[width=\textwidth]{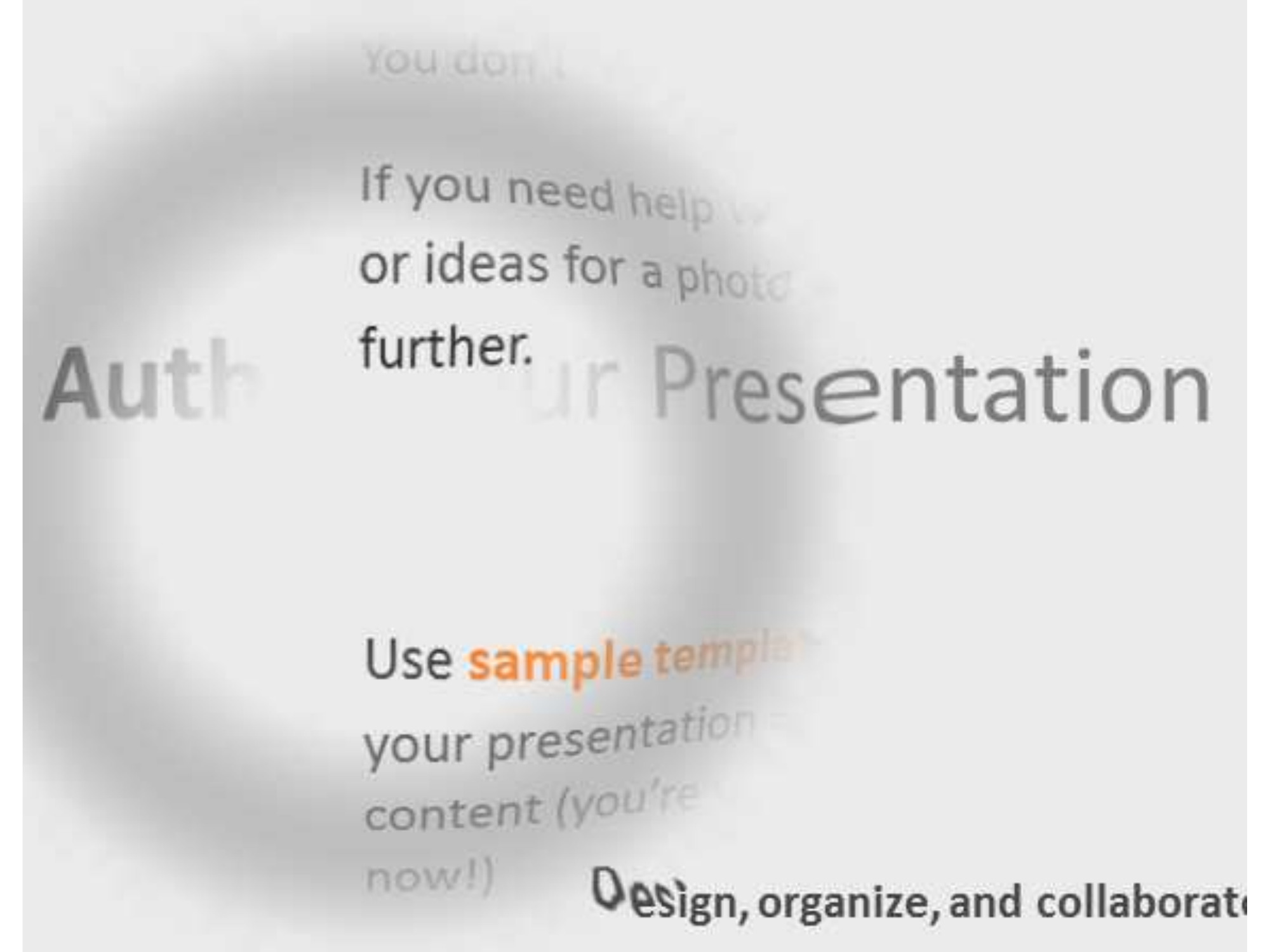}
                \vspace{-0.45cm}
            \hspace{-5cm} 
        \end{subfigure}%
        \begin{subfigure}[b]{0.18\textwidth}
                \includegraphics[width=\textwidth]{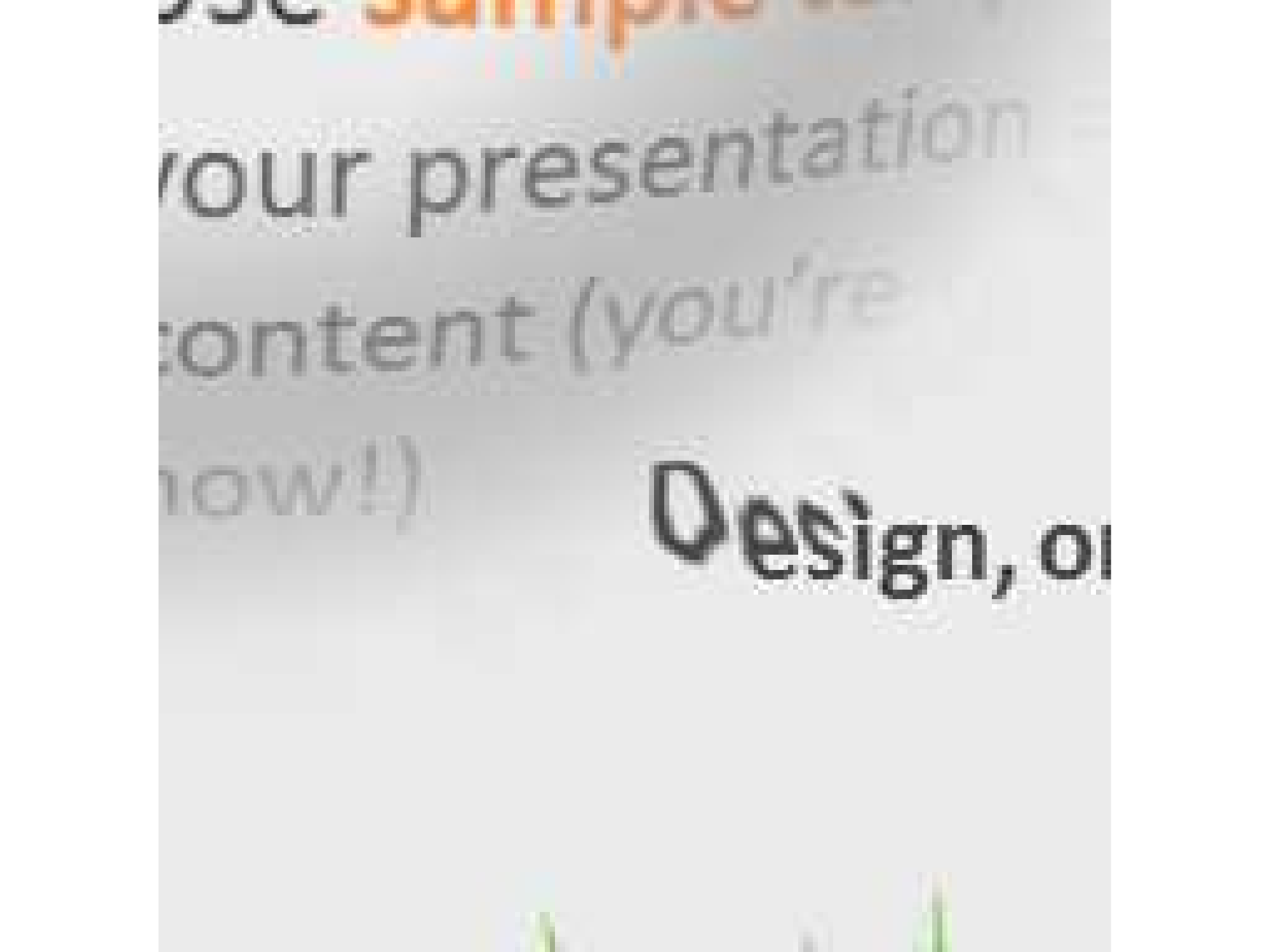}
                 \vspace{-0.45cm}
              \hspace{-4.8cm}
        \end{subfigure}
         \\[1ex]
        \begin{subfigure}[b]{0.18\textwidth}
                \includegraphics[width=\textwidth]{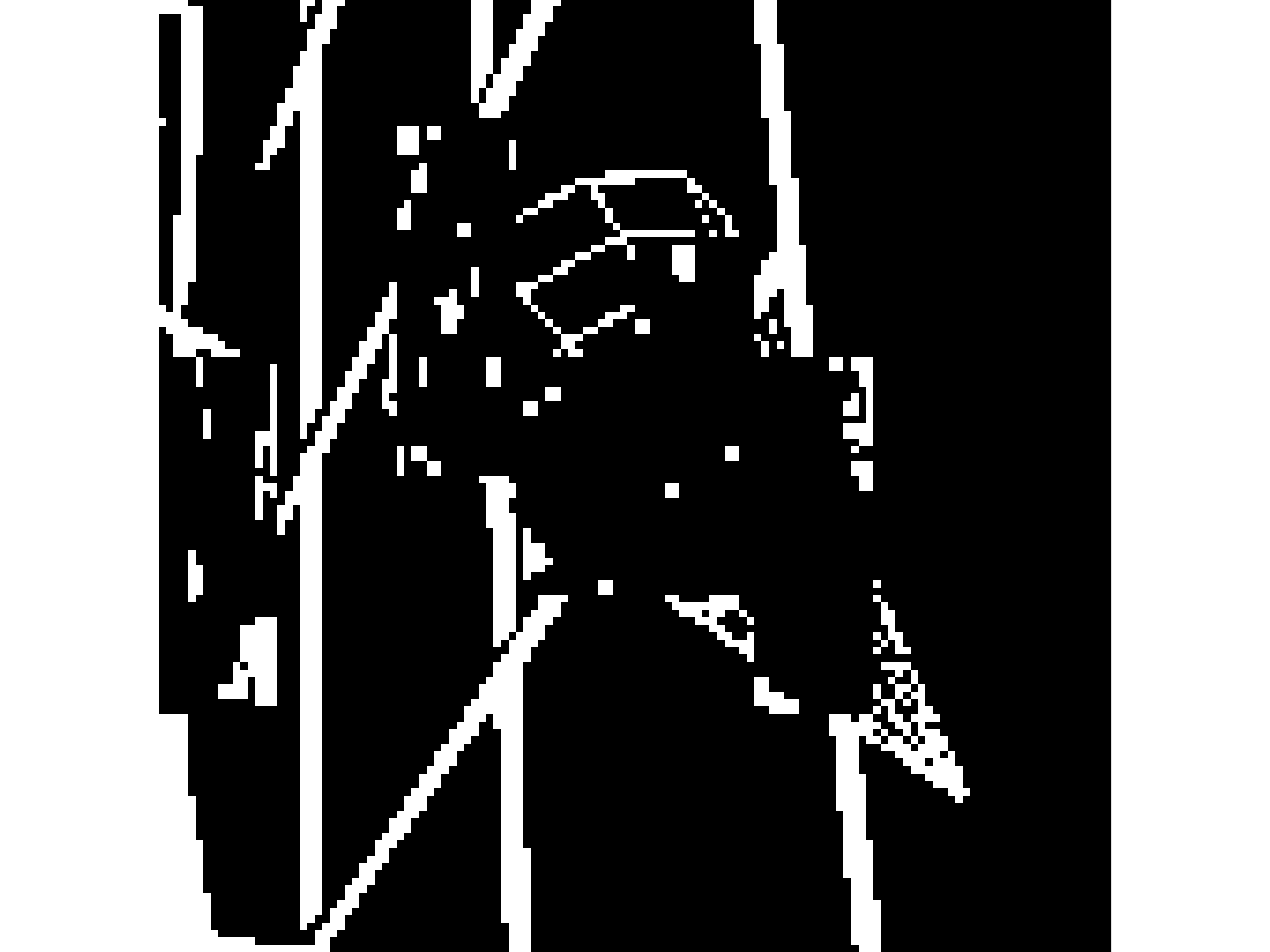}
                                \vspace{-0.5cm}
          \hspace{-2.5cm}    
        \end{subfigure}%
        ~ 
        \begin{subfigure}[b]{0.18\textwidth}
                \includegraphics[width=\textwidth]{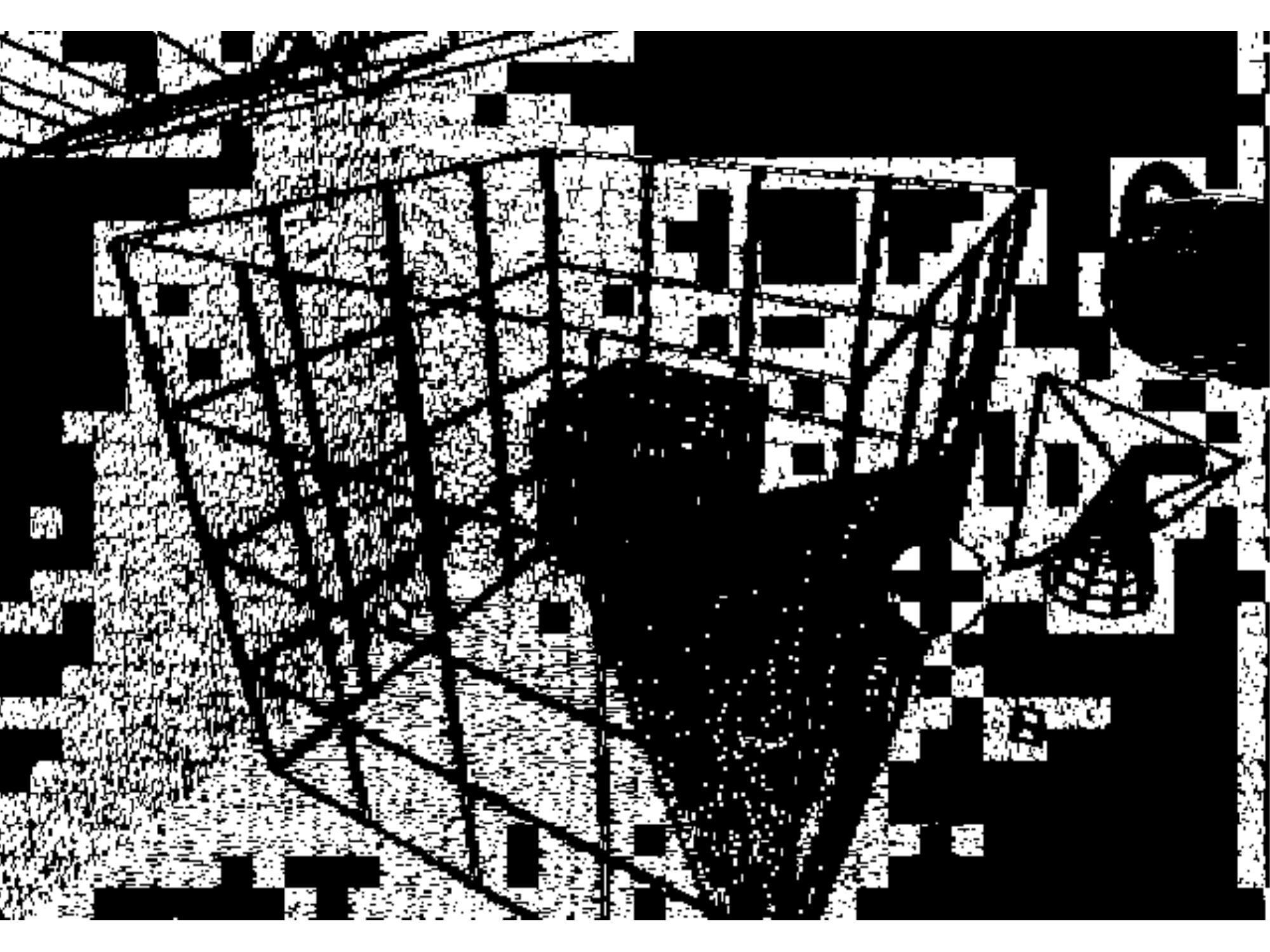}
                \vspace{-0.5cm}
            \hspace{-3cm} 
        \end{subfigure}%
        ~ 
        \begin{subfigure}[b]{0.18\textwidth}
                \includegraphics[width=\textwidth]{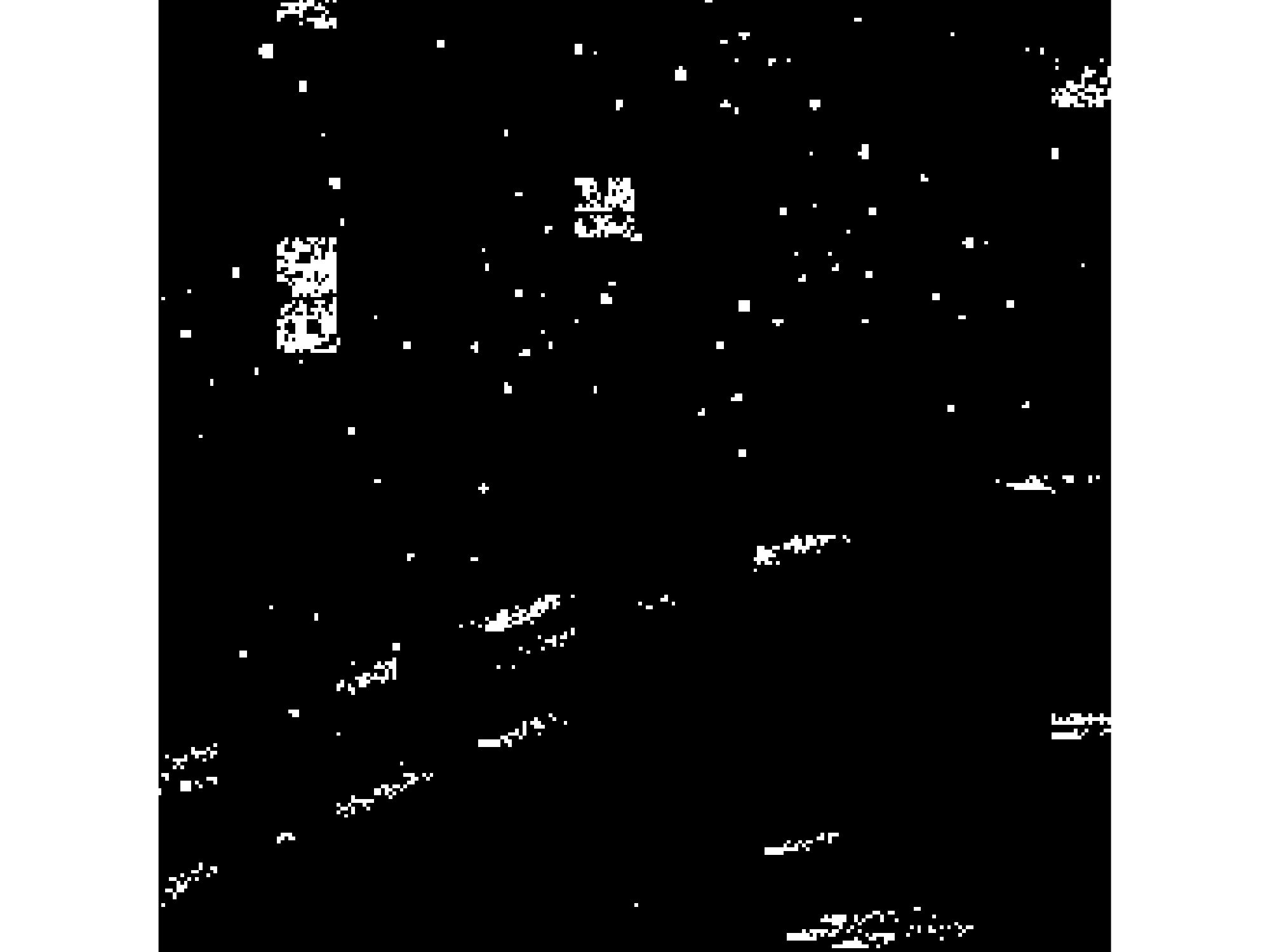}
                \vspace{-0.45cm}
            \hspace{-3cm} 
        \end{subfigure}%
        \begin{subfigure}[b]{0.18\textwidth}
			~ 
                \includegraphics[width=\textwidth]{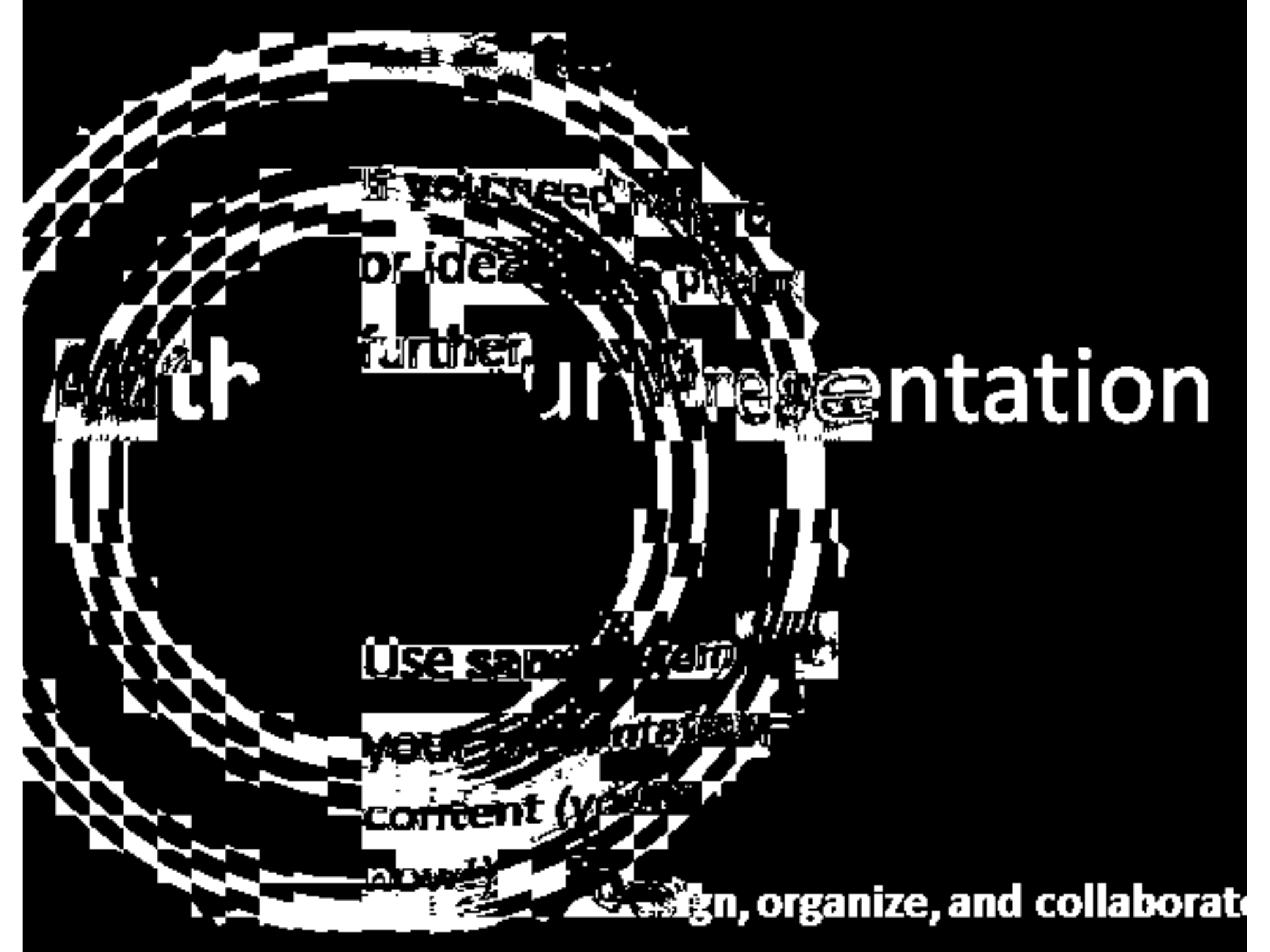}
                \vspace{-0.45cm}
            \hspace{-3cm} 
        \end{subfigure}%
        \begin{subfigure}[b]{0.18\textwidth}
                \includegraphics[width=\textwidth]{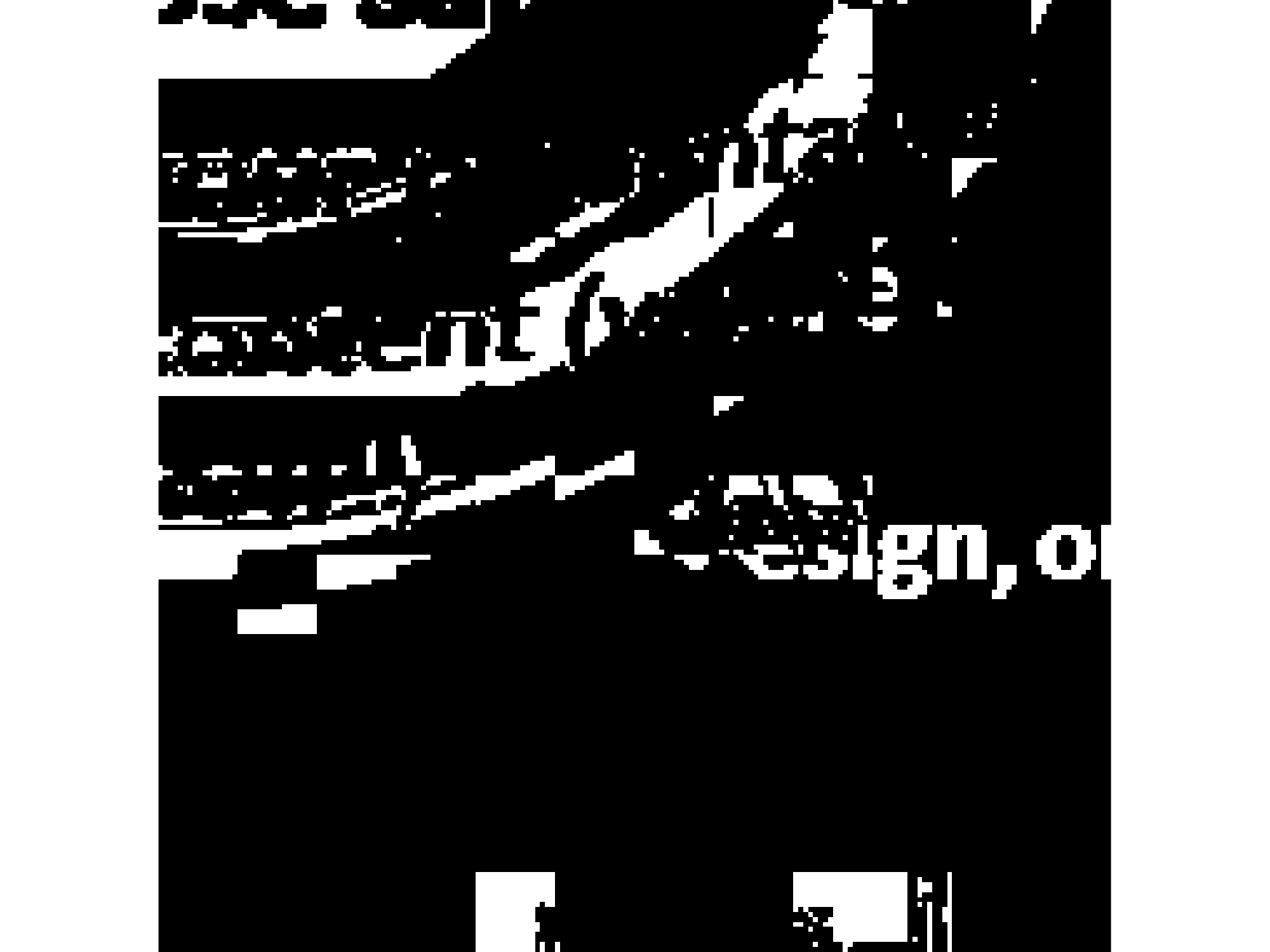}
                 \vspace{-0.45cm}
              \hspace{-4.8cm}
        \end{subfigure} \\[1ex]
        \begin{subfigure}[b]{0.18\textwidth}
                \includegraphics[width=\textwidth]{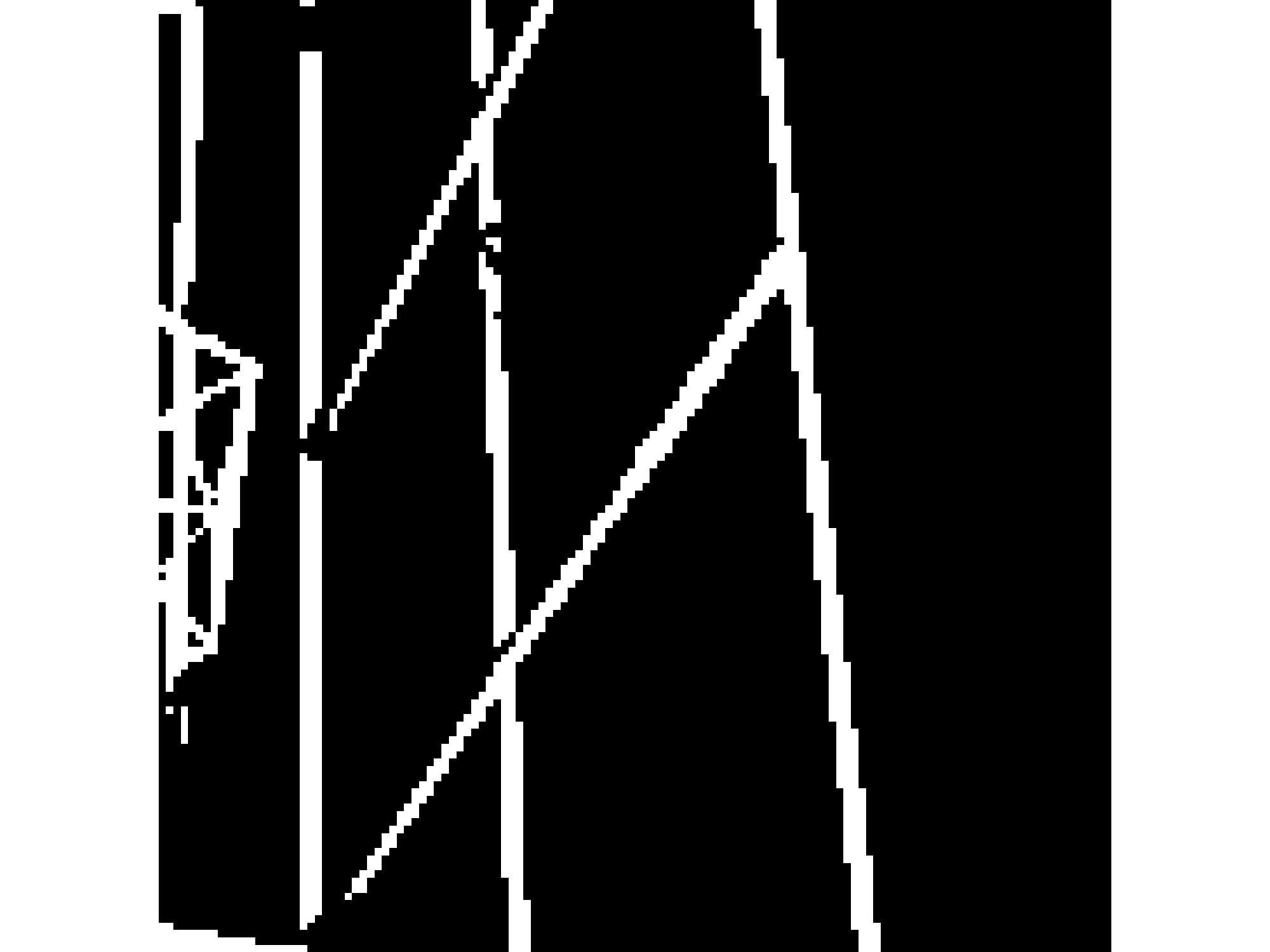}
                                \vspace{-0.5cm}
          \hspace{-2.5cm}    
        \end{subfigure}%
        ~ 
        \begin{subfigure}[b]{0.18\textwidth}
                \includegraphics[width=\textwidth]{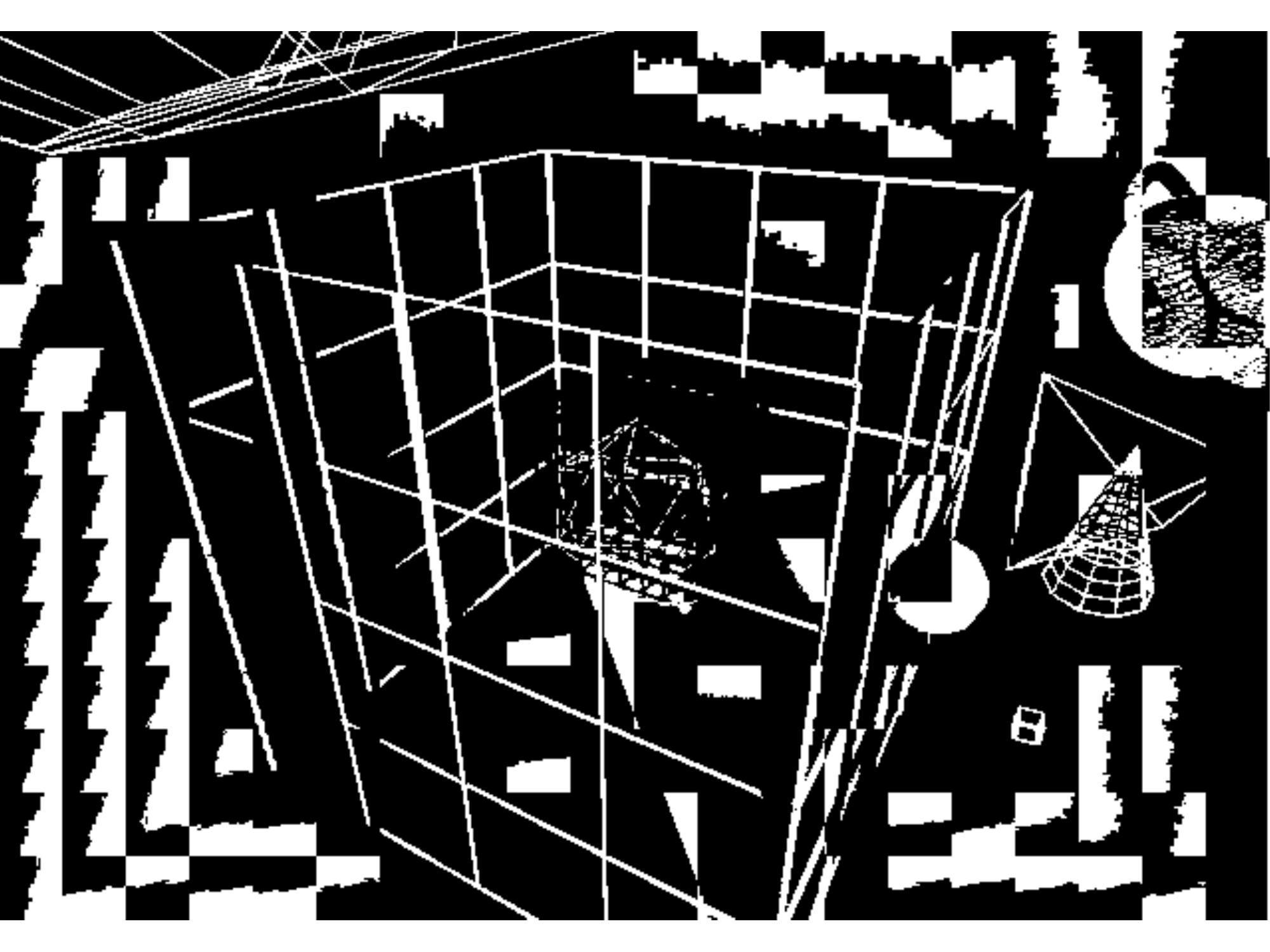}
                \vspace{-0.5cm}
            \hspace{-3cm} 
        \end{subfigure}%
        ~ 
        \begin{subfigure}[b]{0.18\textwidth}
                \includegraphics[width=\textwidth]{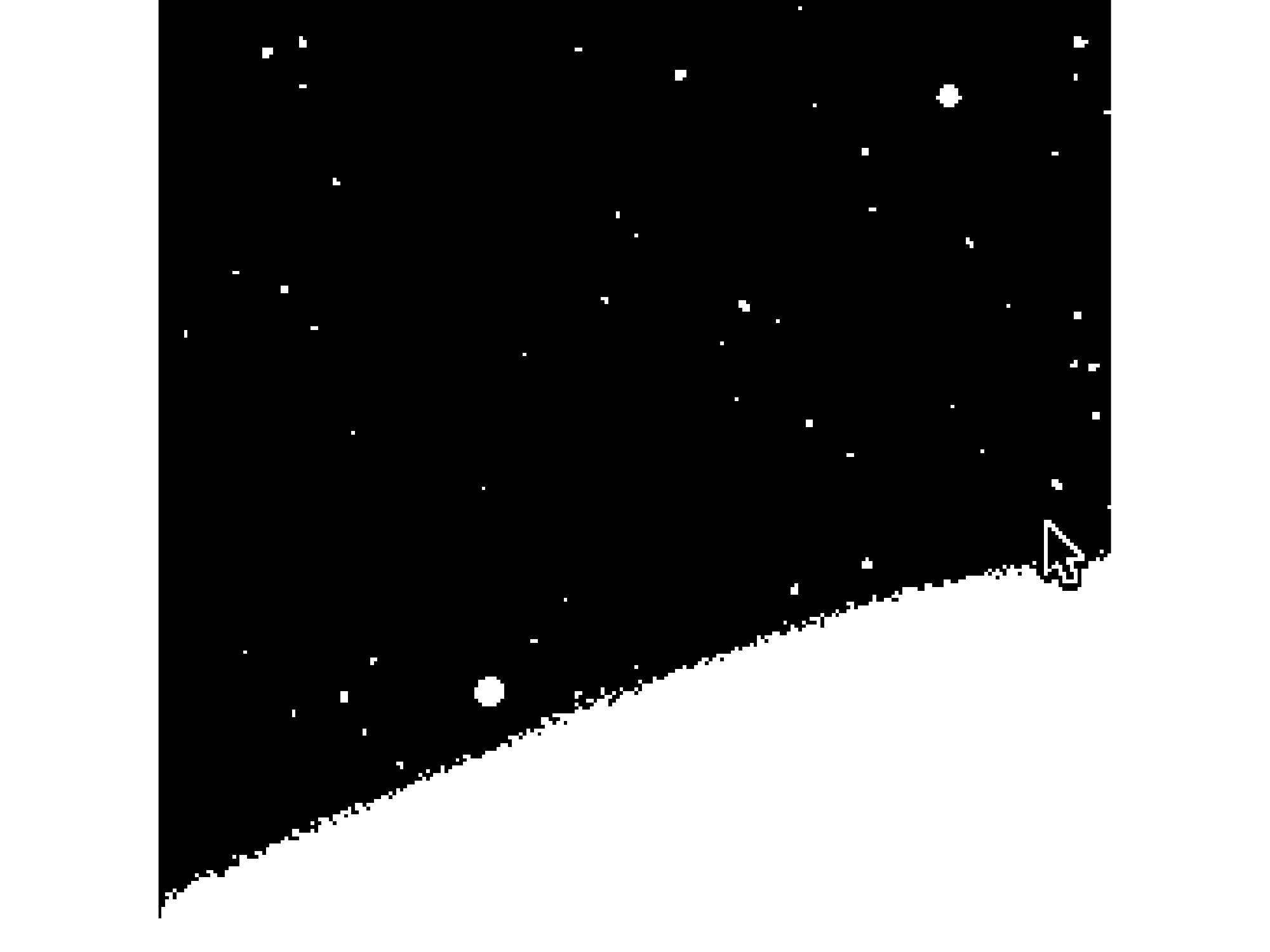}
                \vspace{-0.45cm}
            \hspace{-3cm} 
        \end{subfigure}%
        \begin{subfigure}[b]{0.18\textwidth}
			~ 
                \includegraphics[width=\textwidth]{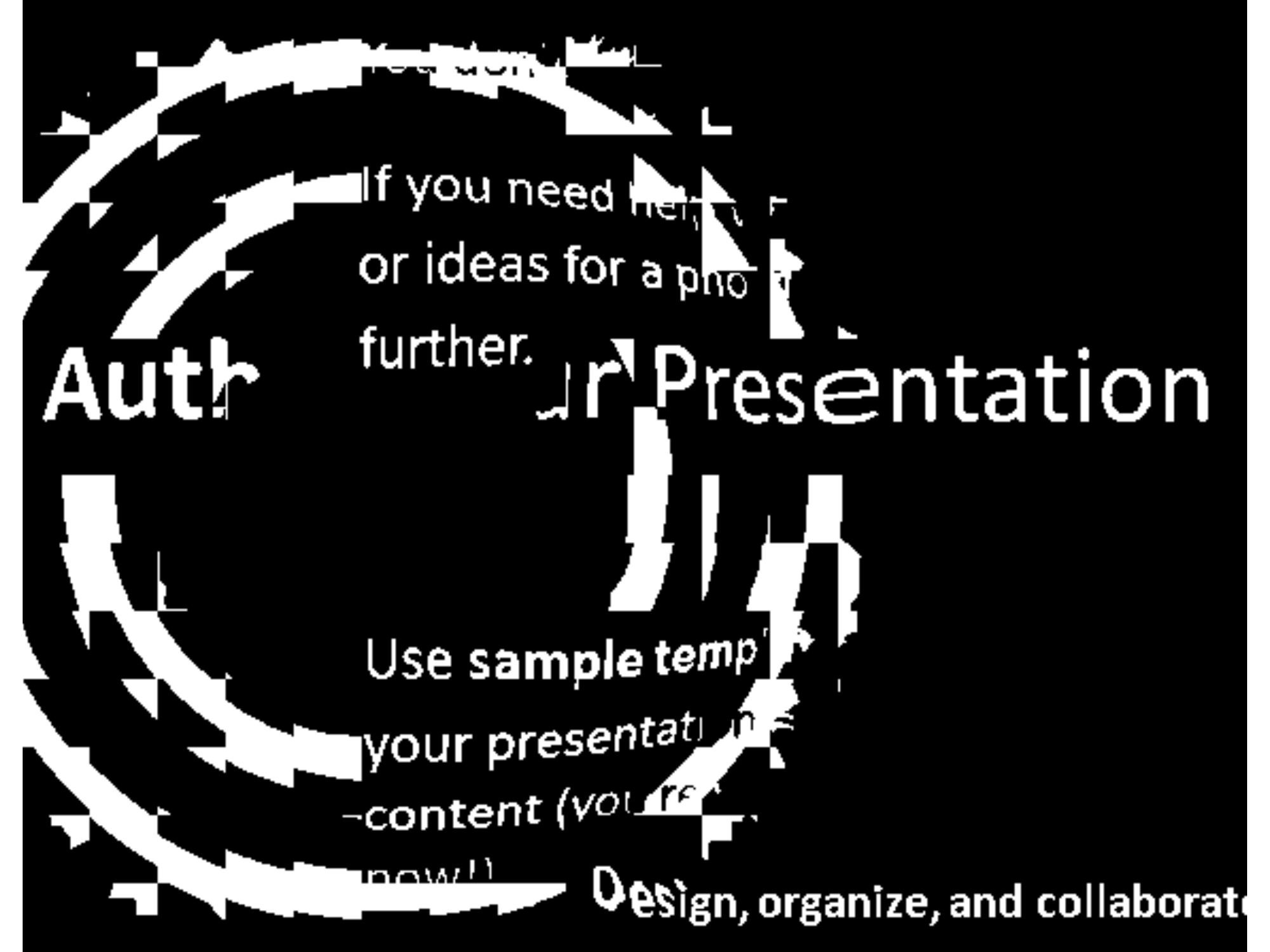}
                \vspace{-0.45cm}
            \hspace{-3cm} 
        \end{subfigure}%
        \begin{subfigure}[b]{0.18\textwidth}
                \includegraphics[width=\textwidth]{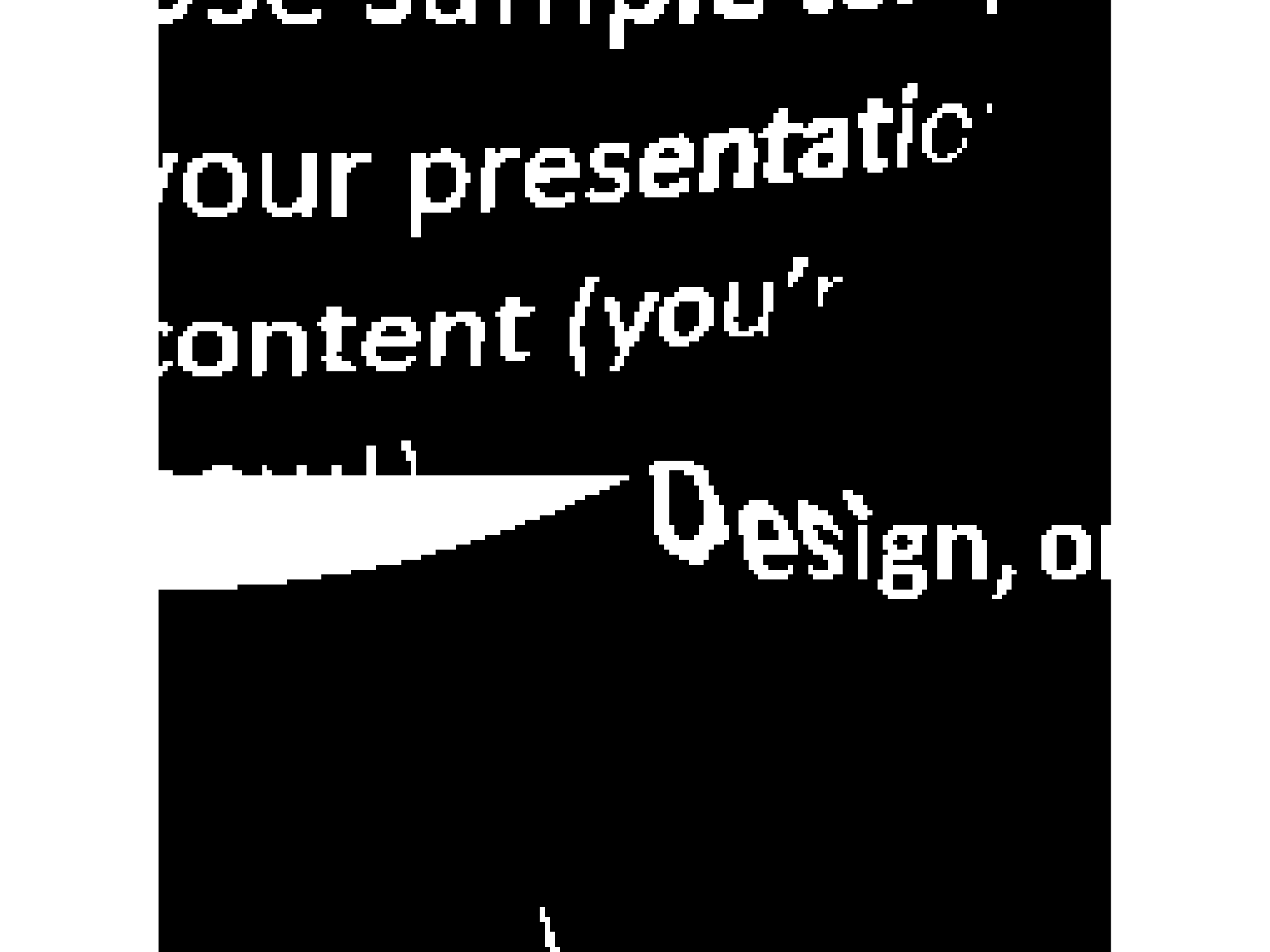}
                 \vspace{-0.45cm}
              \hspace{-4.8cm}
        \end{subfigure}\\[1ex]        
        \begin{subfigure}[b]{0.18\textwidth}
                \includegraphics[width=\textwidth]{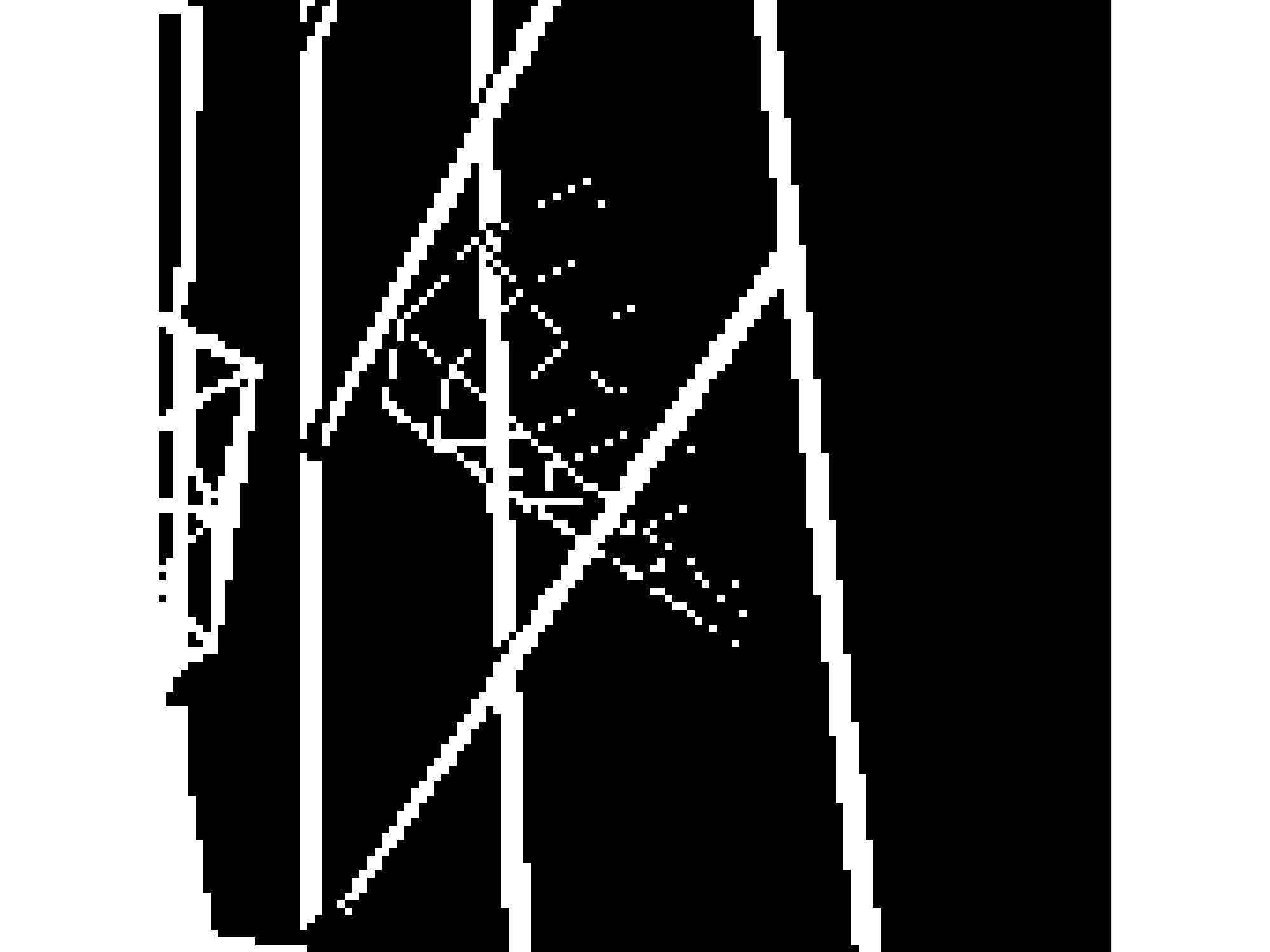}
                                \vspace{-0.5cm}
          \hspace{-2.5cm}    
        \end{subfigure}%
        ~ 
        \begin{subfigure}[b]{0.18\textwidth}
                \includegraphics[width=\textwidth]{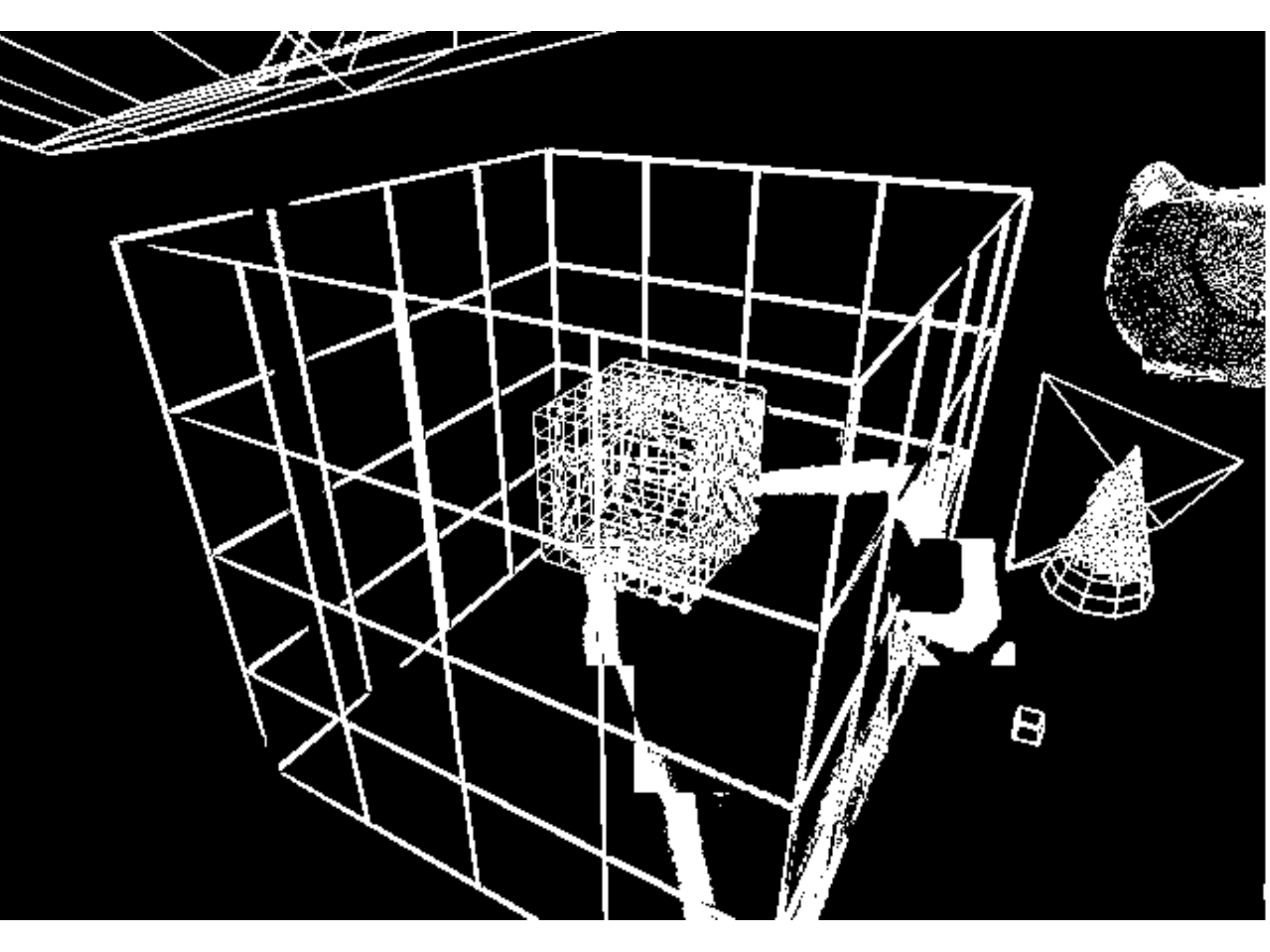}
                \vspace{-0.5cm}
            \hspace{-3cm} 
        \end{subfigure}%
        ~ 
        \begin{subfigure}[b]{0.18\textwidth}
                \includegraphics[width=\textwidth]{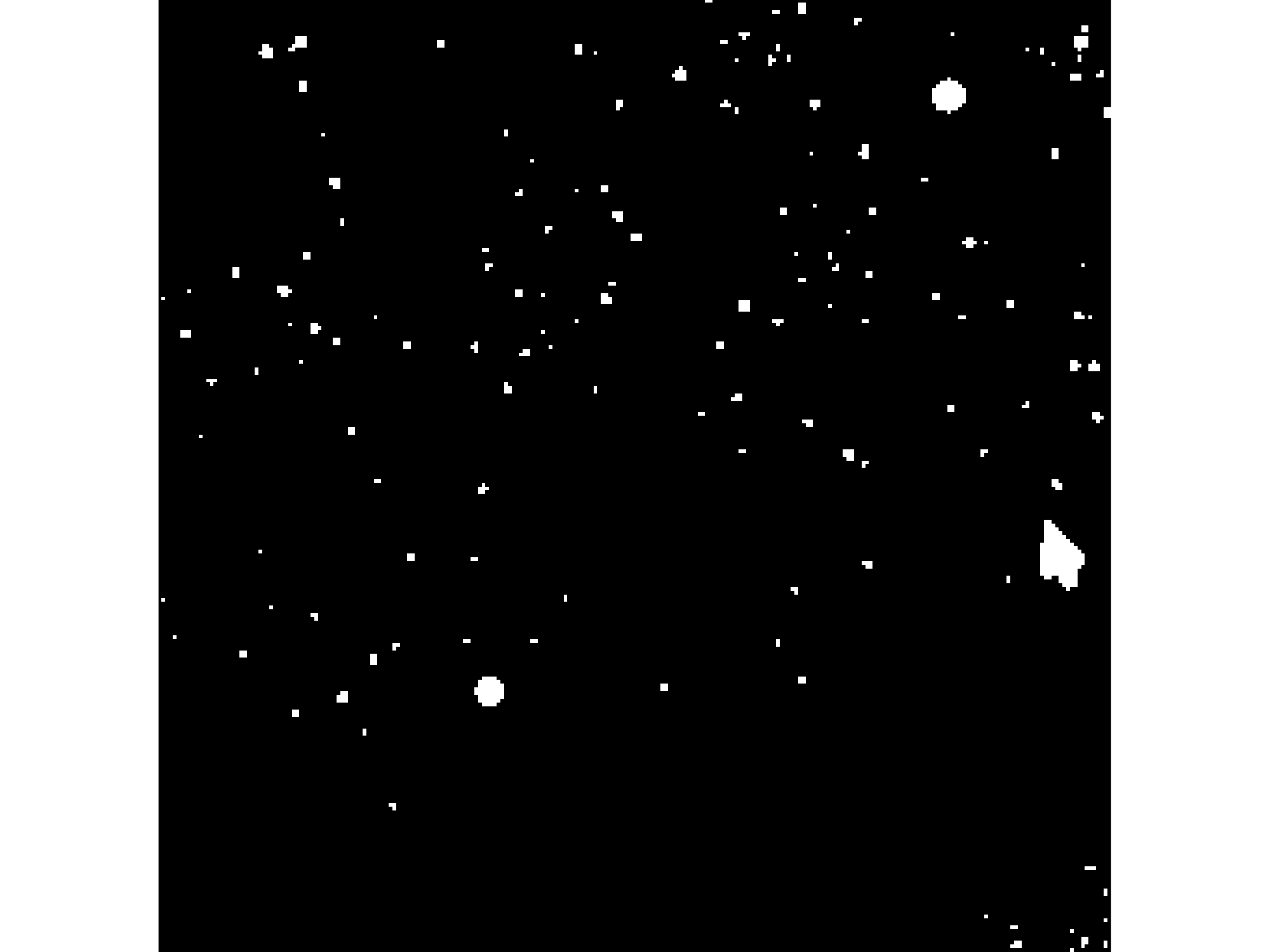}
                \vspace{-0.45cm}
            \hspace{-3cm} 
        \end{subfigure}%
        \begin{subfigure}[b]{0.18\textwidth}
			~ 
                \includegraphics[width=\textwidth]{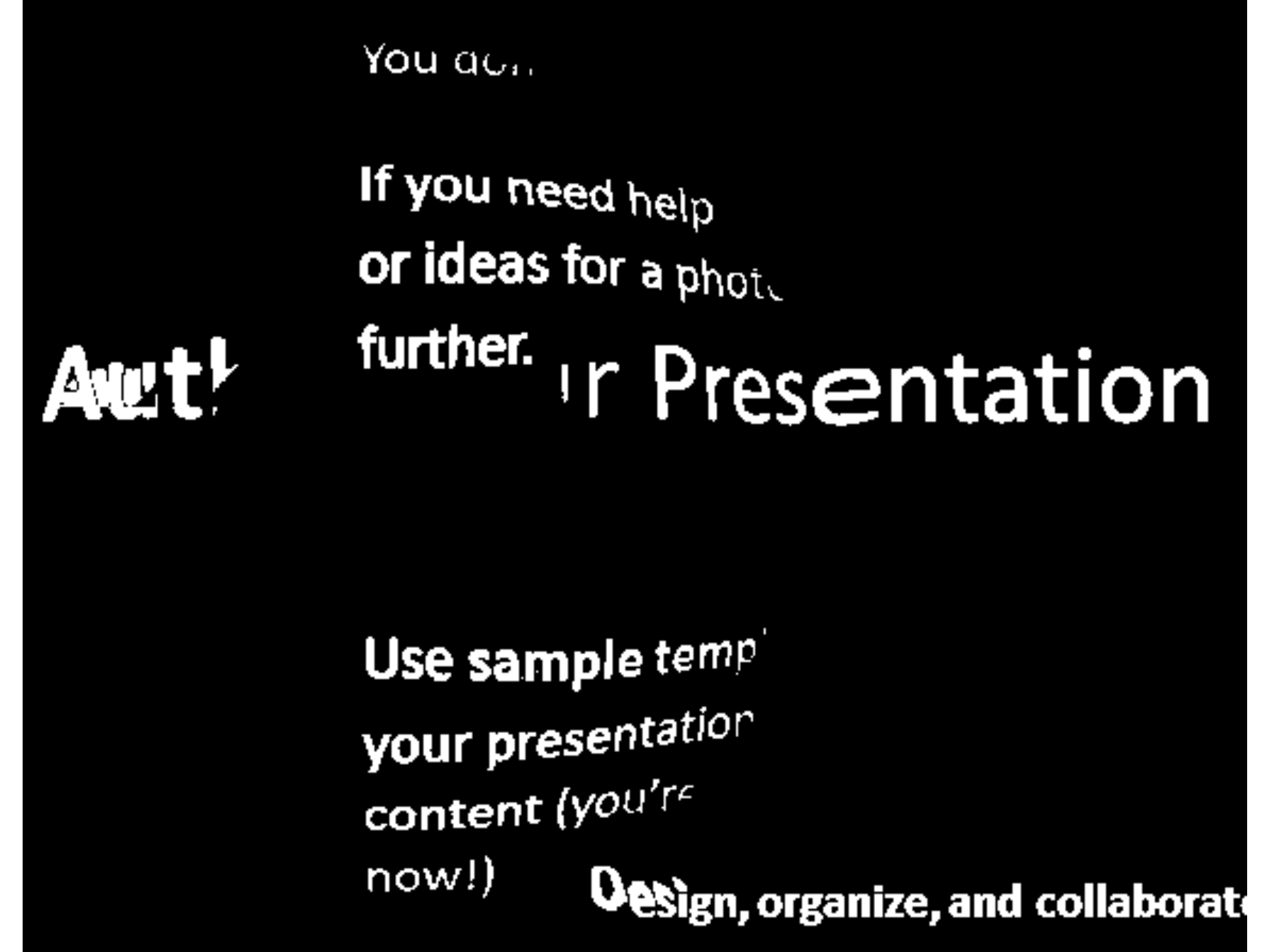}
                \vspace{-0.45cm}
            \hspace{-3cm} 
        \end{subfigure}%
        \begin{subfigure}[b]{0.18\textwidth}
                \includegraphics[width=\textwidth]{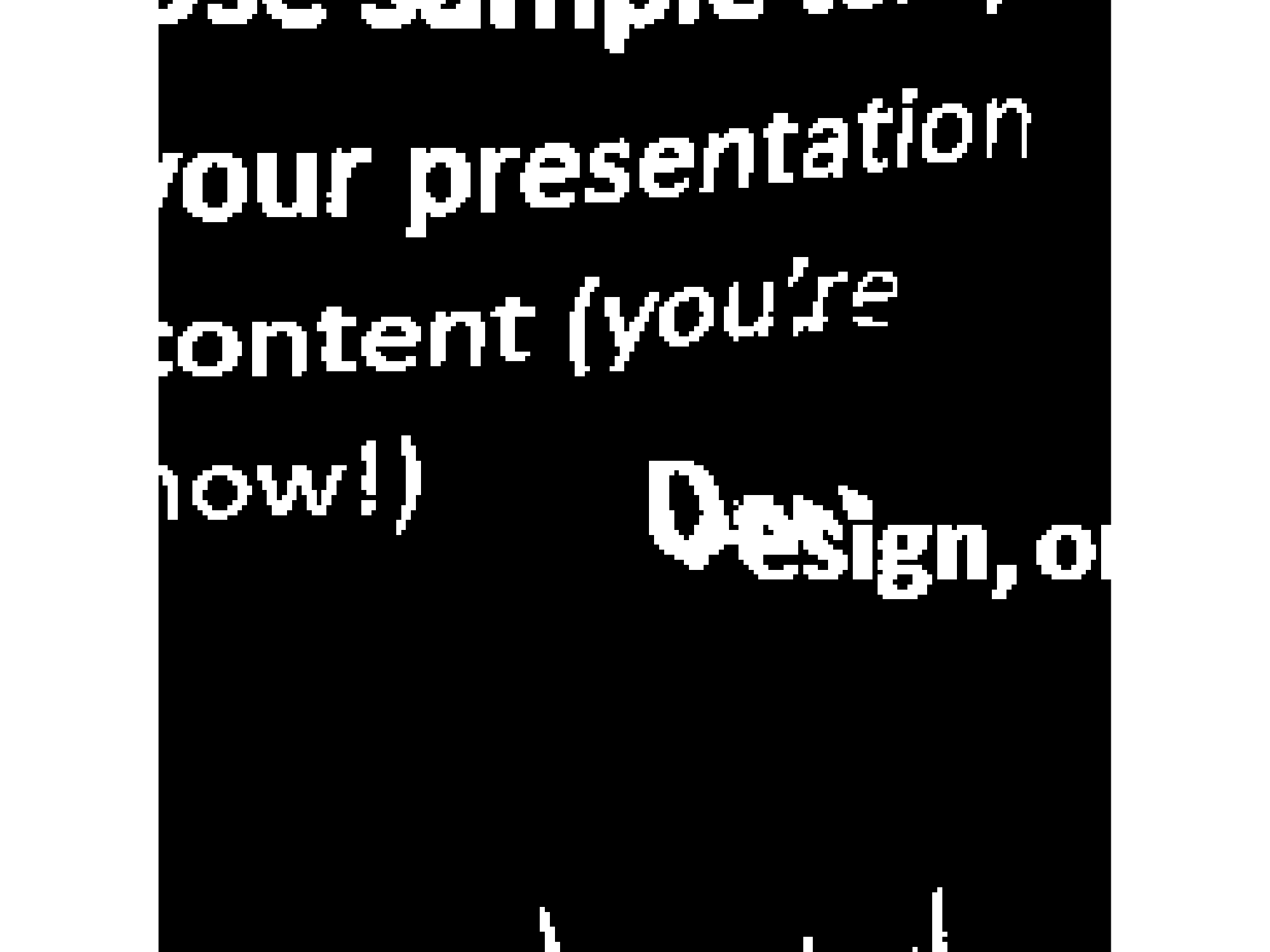}
                 \vspace{-0.45cm}
              \hspace{-4.8cm}
        \end{subfigure} \\[1ex]      
        \begin{subfigure}[b]{0.18\textwidth}
                \includegraphics[width=\textwidth]{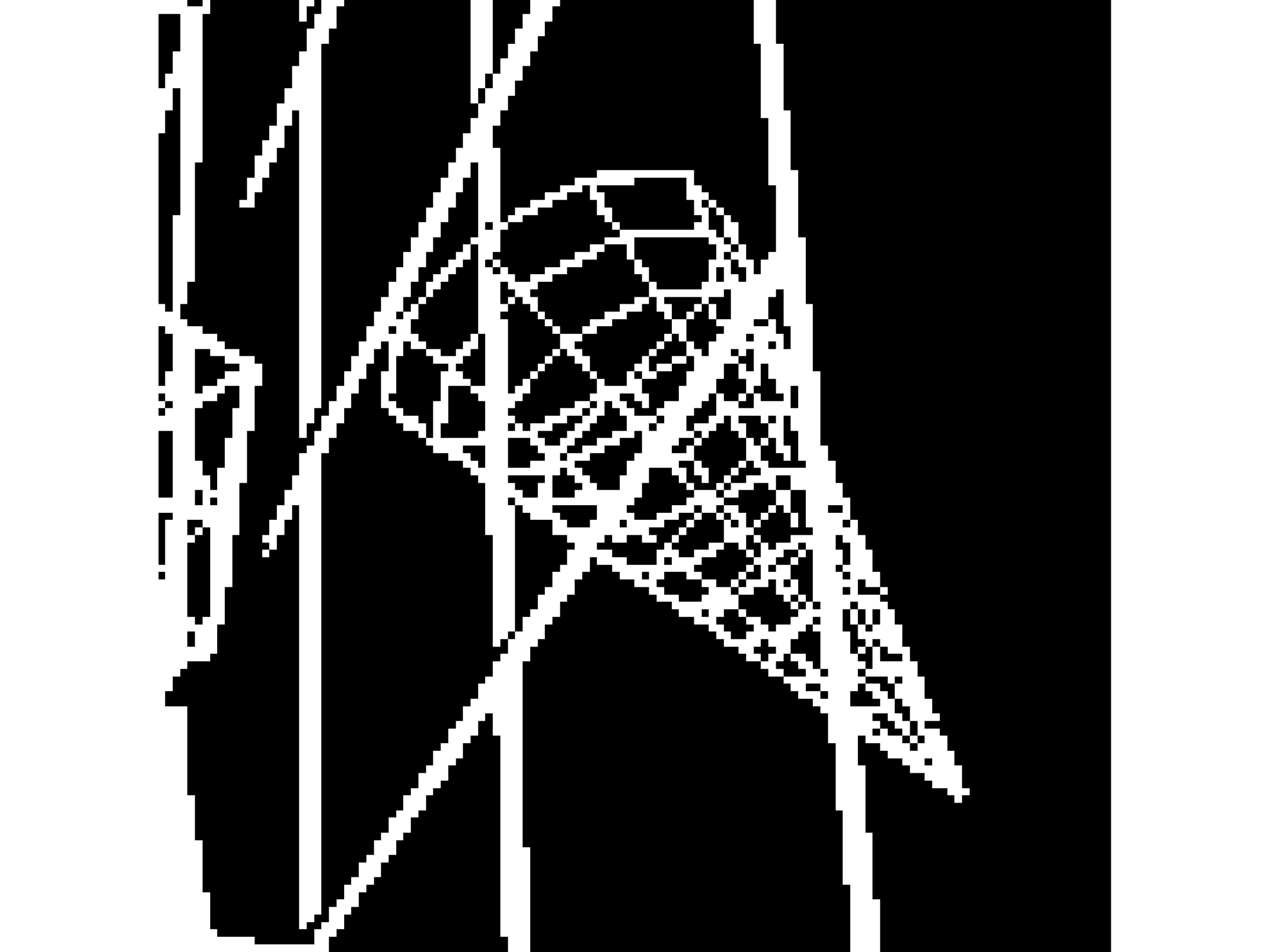}
                                \vspace{-0.5cm}
          \hspace{-2.5cm}    
        \end{subfigure}%
        ~ 
        \begin{subfigure}[b]{0.18\textwidth}
                \includegraphics[width=\textwidth]{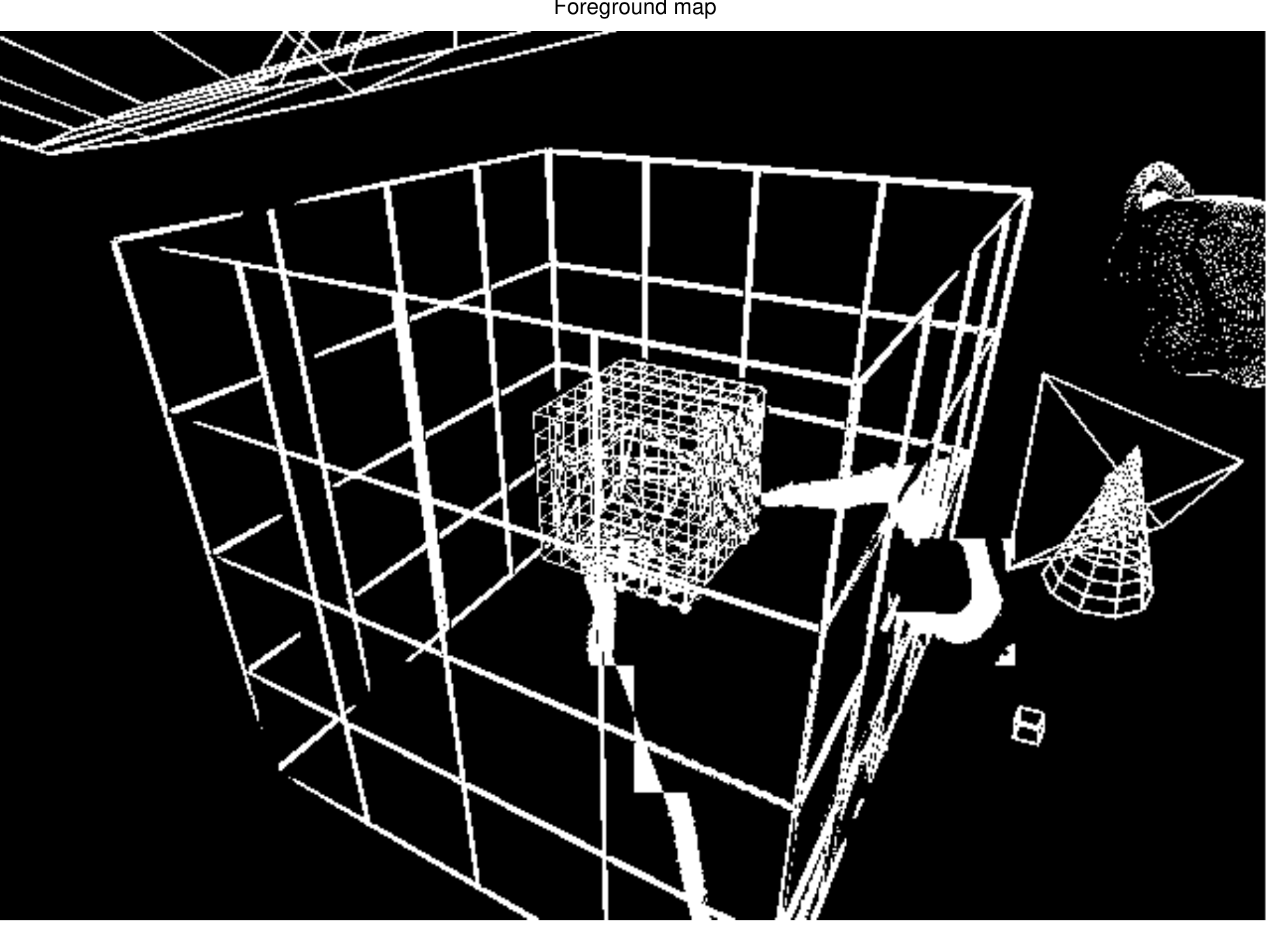}
                \vspace{-0.5cm}
            \hspace{-3cm} 
        \end{subfigure}%
        ~ 
        \begin{subfigure}[b]{0.18\textwidth}
                \includegraphics[width=\textwidth]{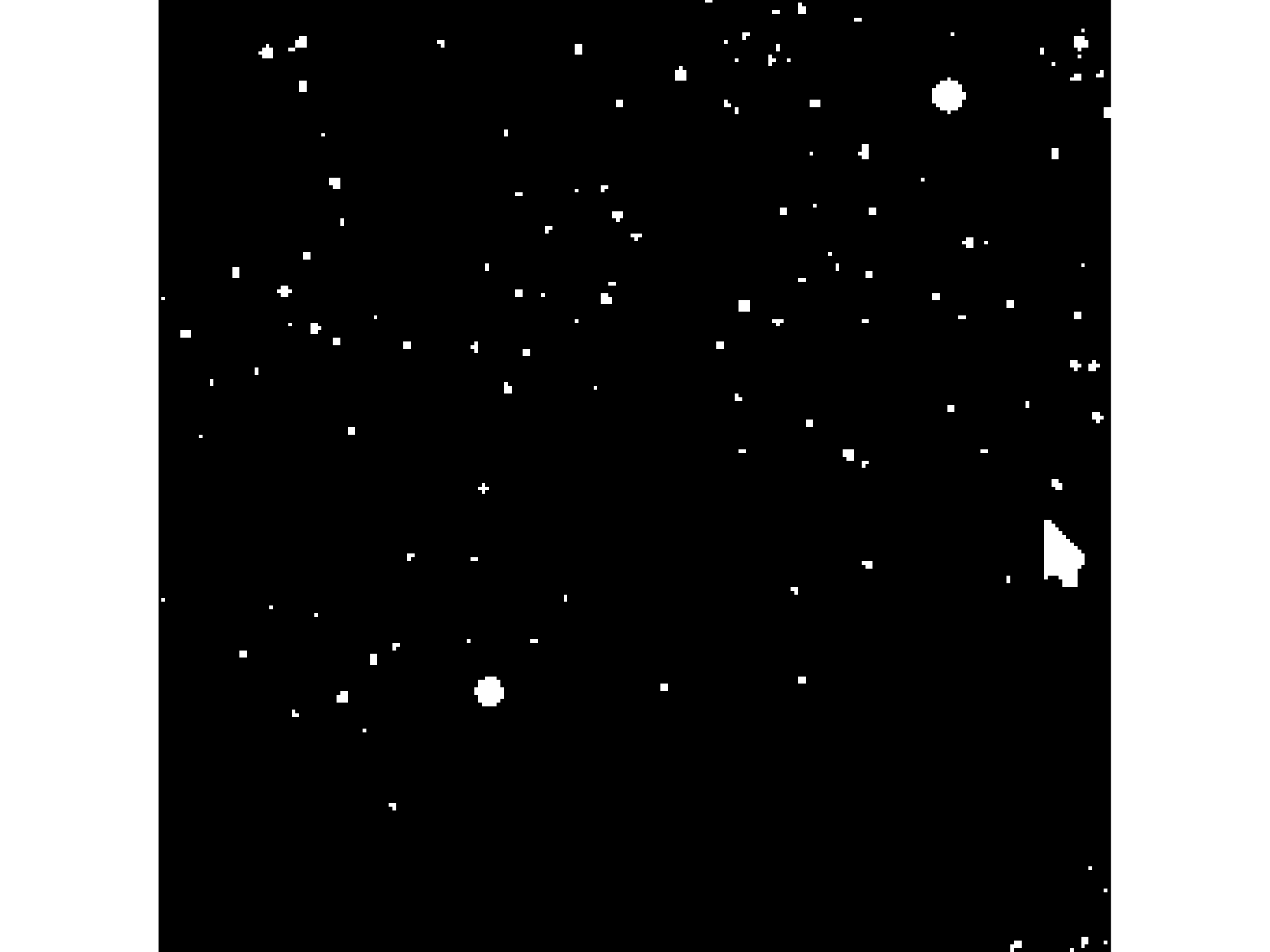}
                \vspace{-0.45cm}
            \hspace{-3cm} 
        \end{subfigure}%
        \begin{subfigure}[b]{0.18\textwidth}
			~ 
                \includegraphics[width=\textwidth]{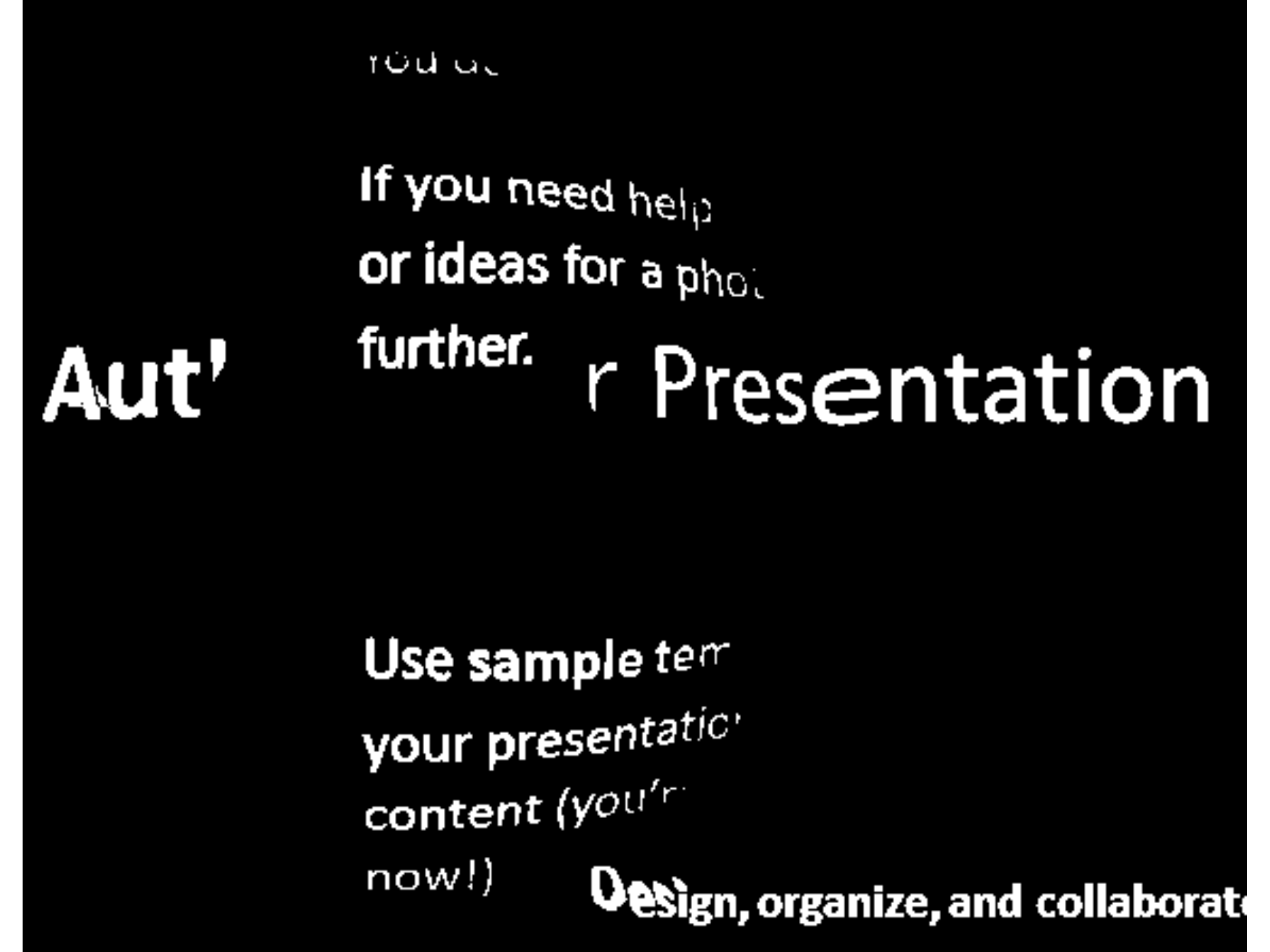} 
                \vspace{-0.45cm}
            \hspace{-3cm} 
        \end{subfigure}%
        \begin{subfigure}[b]{0.18\textwidth}
                \includegraphics[width=\textwidth]{12_GS-eps-converted-to.pdf}
                 \vspace{-0.45cm}
              \hspace{-4.8cm}
        \end{subfigure}
        \caption{Segmentation result for the selected test images. The images in the first row denotes the original images. And the images in the second, third, forth and the fifth rows denote the foreground map by shape primitive extraction and coding, hierarchical k-means clustering, least absolute deviation fitting and the proposed algorithm respectively.}
\end{figure*}

As it can be seen, the proposed scheme achieves much higher precision and recall than hierarchical k-means clustering and SPEC algorithms. Compared to the least absolute deviation fitting, the proposed formulation yields significant improvement in terms of precision, while also having a slightly higher recall rate.

To see the visual quality of the segmentation, the results for 5 test images (each consisting of multiple 64$\times$64 blocks) are shown in Fig. 2. It can be seen that the proposed algorithm gives superior performance over DjVu and SPEC in all cases.
There are also noticeable improvement over our prior approach using least absolute deviation fitting.
We would like to note that, this dataset mainly consists of challenging images where the background and foreground have overlapping color ranges. For simpler cases where the background has a narrow color range that is quite different from the foreground, both DjVu and least absolute deviation fitting will work well. On the other hand,  SPEC  does not work well when the background is fairly homogeneous within a block and the foreground text/lines have varying colors.

To illustrate the smoothness of the background layer and its suitability for being coded with transform-based coding, the filled background layer of a sample image is presented in Fig. 3. The background holes (pixels belonging to the foreground layer) are filled by the predicted value using the smooth model, which is obtained using the least squares fitting to the detected background pixels. As we can see, the background layer is very smooth and does not contain any text and graphics.

\begin{figure}[1 h]
\begin{center}
\hspace{-0.1cm}
    \includegraphics [scale=0.21] {2Original_Image-eps-converted-to.pdf}
  \hspace{-0.05cm}  \includegraphics [scale=0.21] {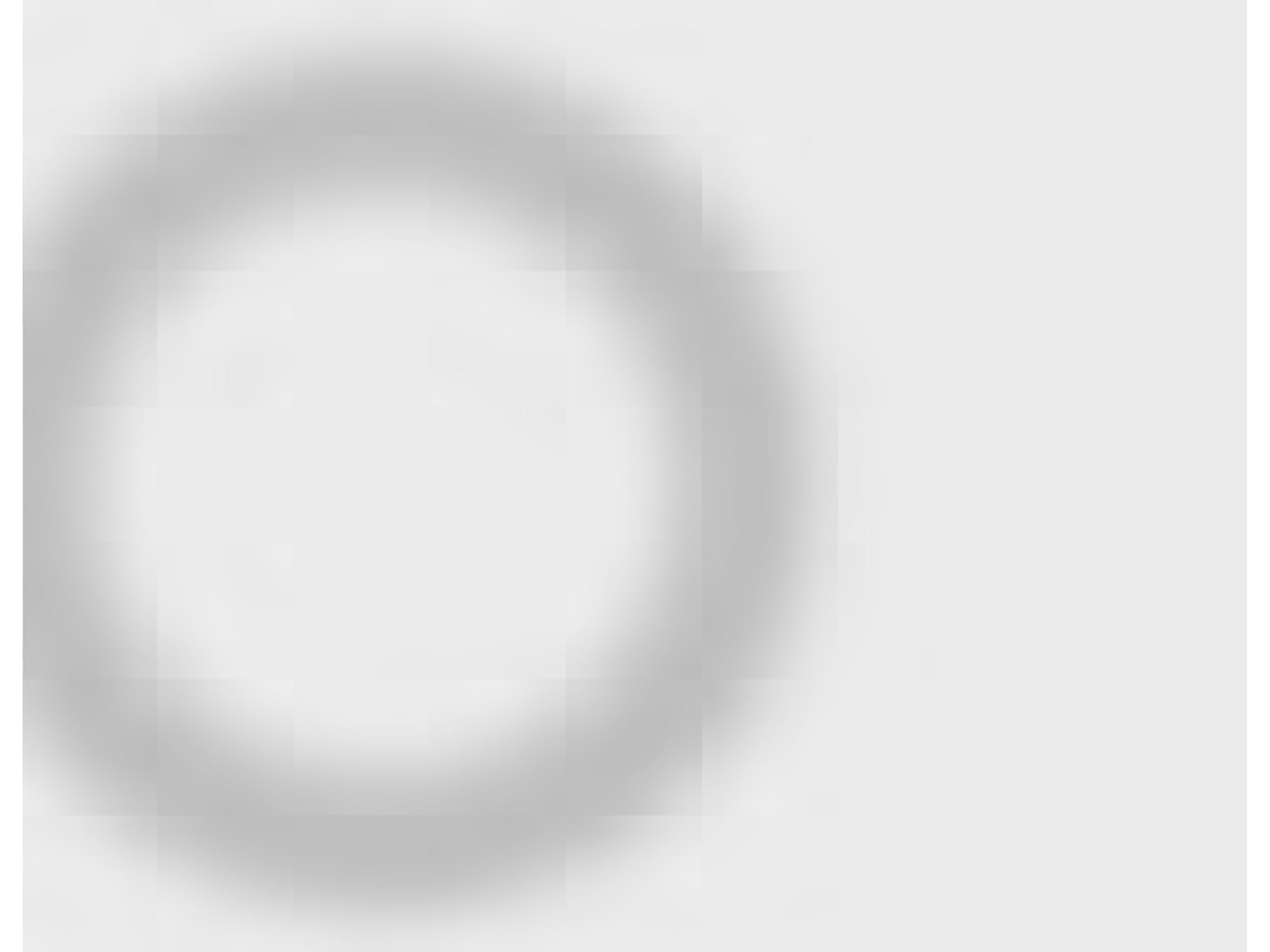}
\end{center}
  \caption{The reconstructed background of an image}
\end{figure}

\section{Conclusion}
This paper proposed a new approach toward foreground/background segmentation in images. 
The background is modeled as a linear combination of a set of low-frequency 2D DCT functions and foreground as a sparse component overlaid on the background. 
We propose a sparse decomposition framework to decompose the image into these two layers.
The group sparsity of foreground is penalized to promote the connectivity of foreground pixels.
Compared to our prior least absolute fitting formulation, the background layer is allowed to choose as many bases from a rich set of smooth functions, but the coefficients are enforced to be sparse so that will not falsely include foreground pixels.
This algorithm has been tested on several test images and compared with some of the previous well-known algorithms for background/foreground separation and has shown significantly better performance in most cases.
In terms of applications, this algorithm can be used in layered coding in video compression, as a pre-processing step in text extraction, and also in medical image segmentation.

\section*{Acknowledgment}
The authors would like to thank Ivan Selesnick and Carlos Fernandez-Granda for their valuable comments and feedback.

\ifCLASSOPTIONcaptionsoff
  \newpage
\fi


\begin{thebibliography}{1}
\begin{small}
\bibitem{MRC}
R.L. DeQueiroz, R.R. Buckley and M. Xu, ``Mixed raster content (MRC) model for compound image compression'', Electronic Imaging'99. International Society for Optics and Photonics, 1998.
\bibitem{omid1}
O Haji-Maghsoudi, A Talebpour, H Soltanian-Zadeh, N maghsoodi, ``Automatic organs detection in WCE'', International Symposium on AI and Signal Processing (AISP), IEEE, 2012.
\bibitem{text}
J. Zhang and R. Kasturi, ``Extraction of Text Objects in Video Documents: Recent Progress'', Document Analysis Systems. 2008.
\bibitem{omid2}
O Maghsoudi, A Talebpour, H Soltanian-Zadeh, M Alizadeh, H Soleimani, ``Informative and Uninformative Regions Detection in WCE Frames'', Journal of Advanced Computing 3.1: 12-34, 2014.
\bibitem{text2}
M Zhao, S Li, J Kwok, ``Text detection in images using sparse representation with discriminative dictionaries'', Image and Vision Computing 28.12: 1590-1599, 2010.
\bibitem{omid3}
O Maghsoudi, A Talebpour, H Soltanian, H Soleimani, ``Automatic informative tissue's discriminators in WCE'',  IEEE International Conference on Imaging Systems and Techniques, 2012.
\bibitem{parsa1}
MP  Hosseini,  MR  NazemZadeh,  D  Pompili,  K  Jafari-Khouzani,  K Elisevich,  H  SoltanianZadeh,  ``Comparative  performance  evaluation  of automated  segmentation  methods  of  hippocampus  from  magnetic  resonance images of temporal lobe epilepsy patients'', Medical physics, 43(1), 538-553, 2016.
\bibitem{med}
S Zhang, Y Zhan, DN Metaxas, ``Deformable segmentation via sparse representation and dictionary learning'' Medical Image Analysis 16.7: 1385-1396, 2012.
\bibitem{parsa2}
MP Hosseini, MR Nazem-Zadeh, D Pompili, H Soltanian-Zadeh, ``Statistical  validation  of  automatic  methods  for  hippocampus  segmentation  in MR images of epileptic patients'', International Conference of the IEEE Engineering in Medicine and Biology Society, 2014.
\bibitem{pegah1}
P Faridi, H Danyali, MS Helfroush, MA Jahromi, ``Cancerous Nuclei Detection and Scoring in Breast Cancer Histopathological Images'', arXiv preprint arXiv: 1612.01237, 2016.
\bibitem{pegah2}
P Faridi, H Danyali, M Helfroush and M Jahromi, ``An automatic system for cell nuclei pleomorphism segmentation in histopathological images of breast cancer'', Signal Processing in Medicine and Biology Symposium, IEEE, 2016.
\bibitem{chromosome}
S. Minaee, M. Fotouhi and B.H. Khalaj, ``A geometric approach for fully automatic chromosome segmentation'', IEEE symposium on SPMB, 2014.
\bibitem{djvu}
P. Haffner, P.G. Howard, P. Simard, Y. Bengio and Y. Lecun, ``High quality document image compression with DjVu'', Journal of Electronic Imaging, 7(3), 410-425, 1998.
\bibitem{spec}
T. Lin and P. Hao, ``Compound image compression for real-time computer screen image transmission'', IEEE Transactions on Image Processing, 14(8), 993-1005, 2005.
\bibitem{LAD}
S. Minaee and Y. Wang, ``Screen content image segmentation using least absolute deviation fitting'', IEEE International Conference on Image Processing, pp.3295-3299, Sept. 2015.
\bibitem{SPSD}
S Minaee, A Abdolrashidi and Y Wang, ``Screen Content Image Segmentation Using Sparse-Smooth Decomposition'', Asilomar Conference on Signals,Systems, and Computers, IEEE, 2015.
\bibitem{wright}
J Wang, C Lu, M Wang, P Li, S Yan and X Hu, ``Robust face recognition via adaptive sparse representation'', IEEE Transactions on Cybernetics, 44, no. 12: 2368-2378, 2014.
\bibitem{yang}
C Fernandez-Granda, E Candes, ``Super-resolution via transform-invariant group-sparse regularization'', In Proceedings of the IEEE International Conference on Computer Vision, pp. 3336-3343. 2013.
\bibitem{soltani}
M. Soltani and C. Hegde, ``Demixing sparse signals from nonlinear observations,'' In  Asilomar Conference on Signals,Systems, and Computers, IEEE, 2016.
\bibitem{donoho}
JL. Starck, M. Elad, and DL. Donoho, ``Image decomposition via the combination of sparse representations and a variational approach,'' IEEE Transactions on Image Processing, 14.10: 1570-1582, 2005.
\bibitem{mairal}
J. Mairal, M. Elad, and G. Sapiro, ``Sparse representation for color image restoration'', Image Processing, IEEE Transactions on 17.1: 53-69, 2008.
\bibitem{randomsub1}
M Rahmani, G Atia, ``Innovation pursuit: A new approach to subspace clustering'', arXiv preprint arXiv:1512.00907, 2015.
\bibitem{randomsub2}
M Rahmani, G Atia, ``A subspace learning approach for high dimensional matrix decomposition with efficient column/row sampling'', International Conference on Machine Learning, 2016.
\bibitem{denoise}
J Ren, J Liu, Z Guo, ``Context-aware sparse decomposition for image denoising and super-resolution,'' IEEE Transactions on Image Processing, pp.1456-1469, 2013.
\bibitem{sp_coding}
J Yang, K Yu, T Huang, ``Supervised translation-invariant sparse coding,'' IEEE Conference on Computer Vision and Pattern Recognition (CVPR), 2010.
\bibitem{sp_coding2}
Taalimi,  A  Rahimpour,  C  Capdevila,  Z  Zhang,  H.  Qi,  ``Robust  coupling in space of sparse codes for multi-view recognition'', International Conference on Image Processing, pp. 3897-3901, IEEE, 2016.
\bibitem{seg_tv}
S Minaee and Y Wang, ``Screen Content Image Segmentation Using Sparse Decomposition and Total Variation Minimization'', International Conference on Image Processing, IEEE, 2016.
\bibitem{seg_journal}
S Minaee and Y Wang, ``Screen Content Image Segmentation Using Robust Regression and Sparse Decomposition'', IEEE Journal on Emerging and Selected Topics in Circuits and Systems, no.99, pp.1-12, 2016.
\bibitem{KLT}
Levey, A., and M. Lindenbaum. ``Sequential Karhunen-Loeve basis extraction and its application to images,'' Image Processing, IEEE Transactions on 9.8: 1371-1374, 2000.
\bibitem{DCT}
A.B. Watson, ``Image compression using the discrete cosine transform'', Mathematica journal 4.1: 81, 1994.
\bibitem{bach1}
F Bach, R Jenatton, J Mairal, G. Obozinski, ``Convex optimization with sparsity-inducing norms'', Optimization for Machine Learning, 2011.
\bibitem{patrick1}
P. L. Combettes and V. R. Wajs, ``Signal recovery by proximal forward-backward splitting,'' Multiscale Modeling and Simulation, vol. 4, no. 4, pp. 1168-1200, November 2005.
\bibitem{ADMM}
S. Boyd, N. Parikh, E. Chu, B. Peleato and J. Eckstein, ``Distributed optimization and statistical learning via the alternating direction method of multipliers'', Foundations and Trends in Machine Learning, 3(1), 1-122, 2011.
\bibitem{SLASSO}
R Jenatton, JY Audibert, F Bach, ``structured variable selection with sparsity-inducing norms,'' The Journal of Machine Learning Research: 2777-2824, 2011.
\bibitem{soft}
D. Donoho, ``De-noising by soft-thresholdingm'' IEEE Transactions on Information Theory, 41.3: 613-627, 1995.
\bibitem{SCC_data}
ISO/IEC JTC 1/SC 29/WG 11 Requirements subgroup, ``Requirements for an extension of HEVC for coding of screen content,'' in MPEG 109 meeting, 2014.
\bibitem{SCC_tran}
T. Zhang, B. Ghanem, S. Liu, C. Xum and B. Yin, ``Screen content coding based on HEVC framework,'' IEEE Transactions on Multimedia, 16, no. 5: 1316-1326, 2014.
\bibitem{our_dataset}
https://sites.google.com/site/shervinminaee/research/image-segmentation
\bibitem{metrics}
DM Powers, ``Evaluation: from precision, recall and F-measure to ROC, informedness, markedness and correlation'', 2011.


\end{small}
\end{thebibliography}
\end{document}